\documentclass[conference]{IEEEtran}
\IEEEoverridecommandlockouts
\usepackage{amssymb}

\usepackage{amsmath,amsfonts}
\usepackage{algorithmic}
\usepackage{mathtools}
\usepackage{graphicx}
\usepackage{times}
\usepackage{subfig}
\usepackage[ruled,vlined,linesnumbered]{algorithm2e}
\SetAlCapNameFnt{\scriptsize}
\SetAlCapFnt{\scriptsize}
\usepackage{bm}
\usepackage{bbm}

\usepackage{xcolor}
\usepackage{svg}
\usepackage[normalem]{ulem}
\usepackage[hidelinks]{hyperref}
\usepackage{arydshln}

\begin{document}

\title{Explain and Improve: LRP-Inference Fine-Tuning for Image Captioning Models}

\author{
    \IEEEauthorblockN{\text{Jiamei Sun}$^1$, \text{Sebastian Lapuschkin}$^2$, \text{Wojciech Samek}$^2$,  \text{Alexander Binder}$^3$
    } \\
    \IEEEauthorblockA{$^1$ \text{Information Systems Technology and Design Pillar, Singapore University of Technology and Design (SUTD), Singapore}\\
    $^2$ \text{Department of Artificial Intelligence, Fraunhofer Heinrich Hertz Institute, Berlin, Germany}\\
    $^3$ \text{Department of Informatics, University of Oslo, Oslo, Norway}
    }
}

\maketitle

\begin{abstract}
This paper analyzes the predictions of image captioning models with attention mechanisms beyond visualizing the attention itself. We develop variants of layer-wise relevance propagation (LRP) and gradient-based explanation methods, tailored to image captioning models with attention mechanisms. We compare the interpretability of attention heatmaps systematically against the explanations provided by explanation methods such as LRP, Grad-CAM, and Guided Grad-CAM. We show that explanation methods provide simultaneously pixel-wise image explanations (supporting and opposing pixels of the input image) and linguistic explanations (supporting and opposing words of the preceding sequence) for each word in the predicted captions. We demonstrate with extensive experiments that explanation methods 1) can reveal additional evidence used by the model to make decisions compared to attention; 2) correlate to object locations with high precision; 3) are helpful to ``debug'' the model, e.g. by analyzing the reasons for hallucinated object words. With the observed properties of explanations, we further design an LRP-inference fine-tuning strategy that reduces the issue of object hallucination in image captioning models, and meanwhile, maintains the sentence fluency. We conduct experiments with two widely used attention mechanisms: the adaptive attention mechanism calculated with the additive attention and the multi-head attention mechanism calculated with the scaled dot product.
\end{abstract}

\section{Introduction}
\label{sec:introduction}
Image captioning is a setup that aims at generating text descriptions from image representations. This task requires a comprehensive understanding of the image content and a well-performing decoder which translates image features into sentences. The combination of a convolutional neural network (CNN) and a recurrent neural network (RNN) is a commonly used structure in image captioning models, with CNN as the image encoder and RNN as the sentence decoder \cite{SHOWTELL:vinyals2015show,DEEPVS:karpathy2015deep,CNNLSTM:soh2016learning}. An established feature of image captioning is the attention mechanism that enables the decoder to focus on a sub-region of the image when predicting the next word of the caption 
\cite{SHOWATTENDTEL:Lxu2015show,ERD:NIPS2016_6167,MSM:you2016image,KNOWING:lu2017knowing,BUTD:Anderson_2018_CVPR,AATAttention:huang2019adaptively, HierarchicalAttention:wang2019hierarchical, TRANSFORMER:vaswani2017attention,AOATransformer:huang2019attention, EntangledTransformer:li2019entangled, MultimodelTransformer:yu2019multimodal,MeshedMemoTransformer:cornia2020meshed}. Attentions are usually visualized as attention heatmaps, indicating which parts of the image are related to the generated words. As such, they are a natural resource to explain the prediction of a word. Furthermore, attention heatmaps are usually considered as the qualitative evaluations of image captioning models in addition to the quantitative evaluation metrics such as BLEU \cite{BLEU:papineni-etal-2002-bleu}, METEOR \cite{METEOR:banerjee2005meteor}, ROUGE-L \cite{ROUGE:lin2004rouge}, CIDEr \cite{CIDEr:Vedantam_2015_CVPR}, SPICE \cite{SPICE:anderson2016spice}.

Attention heatmaps provide a certain level of interpretability for image captioning models since they can reflect the locations of objects. However, the outputs of image captioning models rely on not only the image input but also the previously generated word sequence. Attention heatmaps alone meet difficulties in disentangling the contributions of the image input and the text input.

\begin{figure}
    \centering
    \subfloat[image explanations]{\includegraphics[width=0.5\textwidth]{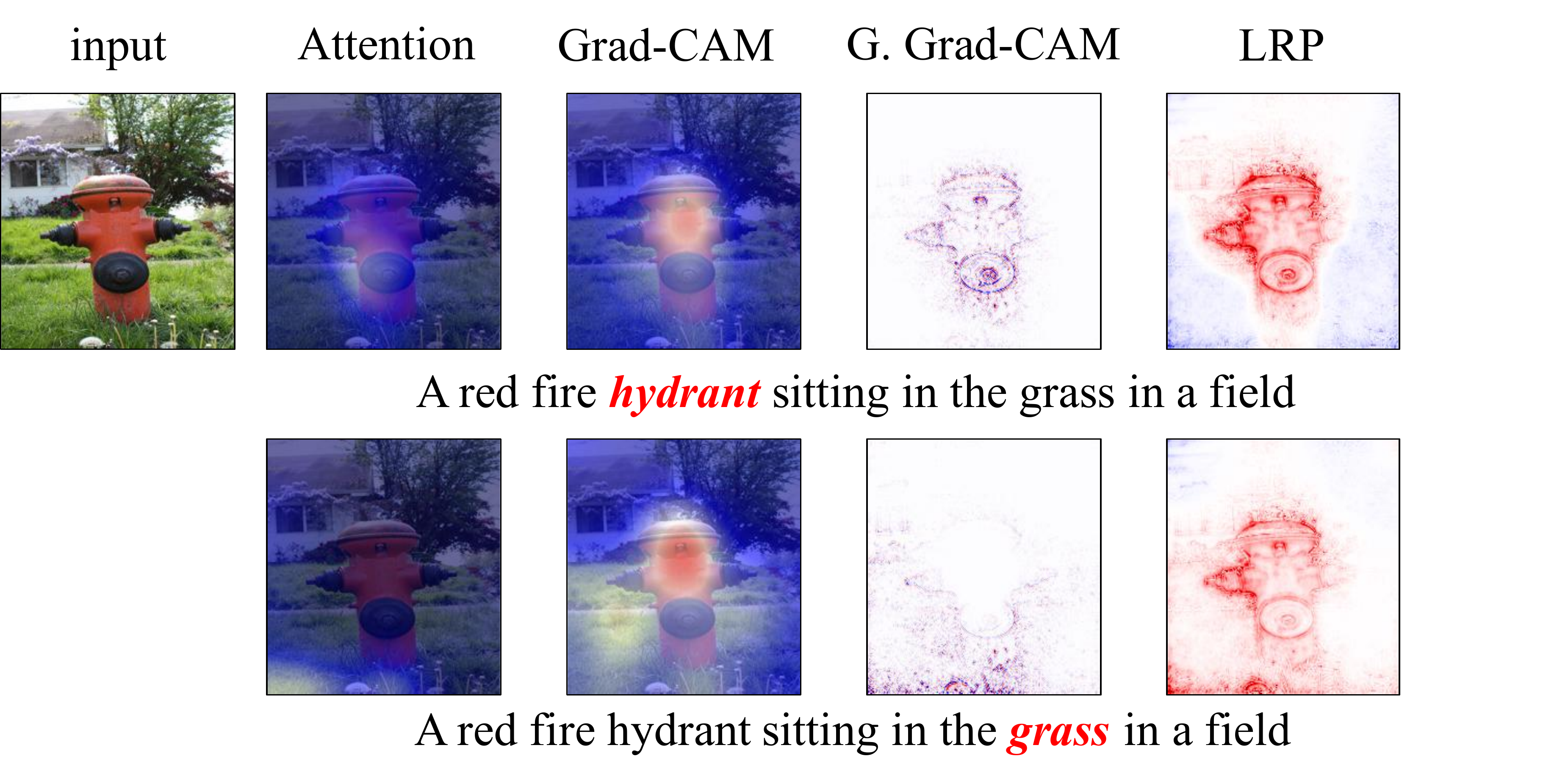}}\\
    \subfloat[linguistic explanations of LRP]{\includegraphics[width=0.49\textwidth]{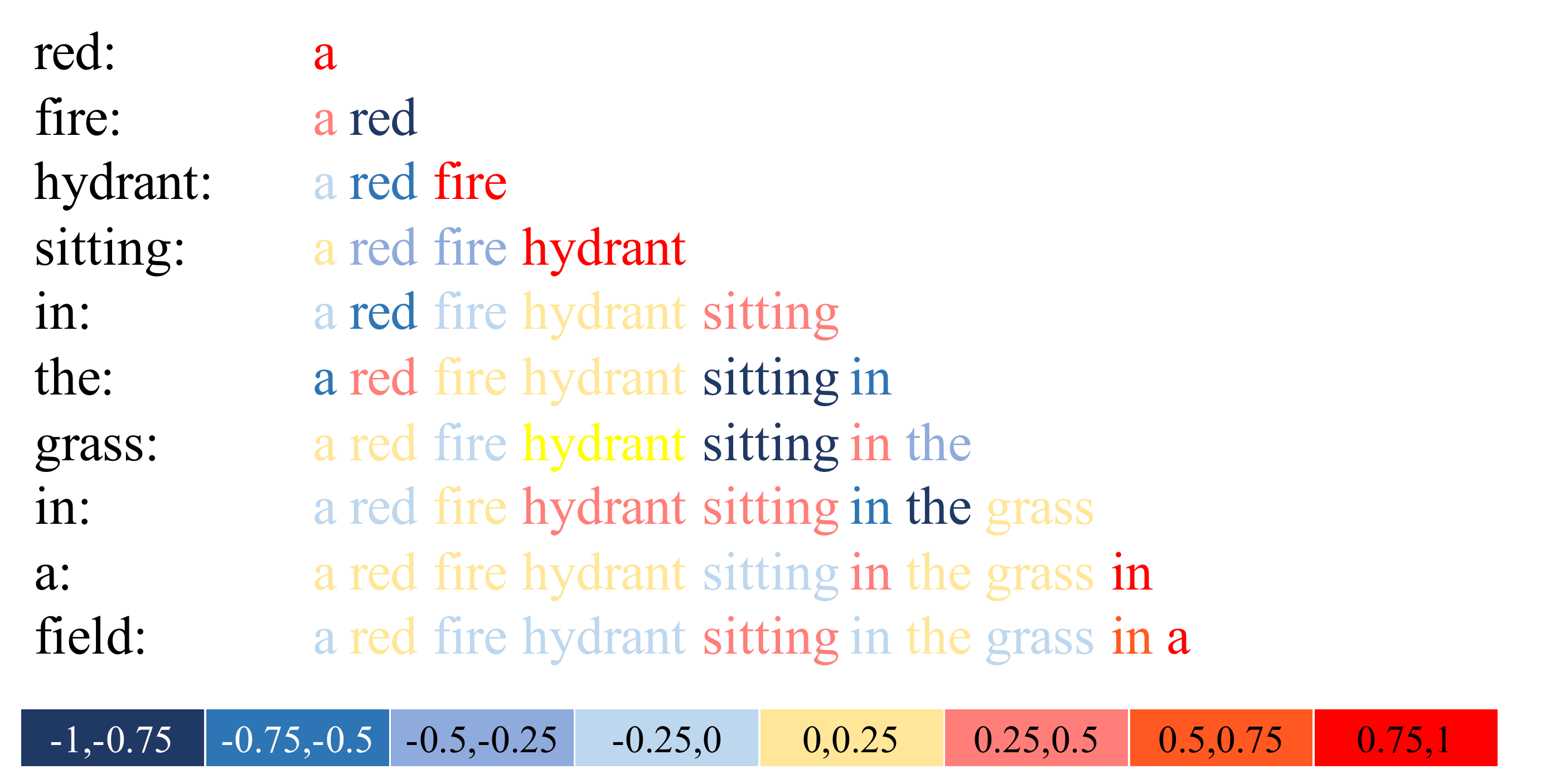}}
    \caption{(a): Image explanations of the word \emph{hydrant} (the first row) and \emph{grass} (the second row) with attention, Grad-CAM, Guided Grad-CAM (G.Grad-CAM) and LRP. (b): The linguistic explanations of LRP for each word in the predicted caption. Blue and red colors indicate negative and positive relevance scores, respectively.
    }
    \label{fig:intro_explanation}
\end{figure}

To gain more insights into the image captioning models, we adapt layer-wise relevance propagation (LRP) and gradient-based explanation methods (Grad-CAM, Guided Grad-CAM \cite{GRADCAM:selvaraju2017grad}, and GuidedBackpropagation \cite{GuidebackPropagation:springenberg2014striving}) to explain image captioning predictions with respect to the image content and the words of the sentence generated so far. These approaches provide high-resolution image explanations for CNN models \cite{GuidebackPropagation:springenberg2014striving,LRP:bach2015pixel}. LRP also provides plausible explanations for LSTM architectures \cite{LRP-LSTM:arras2017explaining,PPAP:sun2019generalized}. Figure \ref{fig:intro_explanation} shows an example of the explanation results of attention-guided image captioning models. 
Taking LRP as an example, both positive and negative evidence is shown in two aspects: 1) for image explanations, the contribution of the image input is visualized as heatmaps; 2) for linguistic explanations, the contribution of the previously generated words to the latest predicted word is shown. 

The explanation results in Figure \ref{fig:intro_explanation} exhibit intuitive correspondence of the explained word to the image content and the related sequential input. However, to our best knowledge, few works quantitatively analyze how accurate the image explanations are grounded to the relevant image content and whether the highlighted inputs are used as evidence by the model to make decisions. We study the two questions by quantifying the grounding property of attention and explanation methods and by designing an ablation experiment for both the image explanations and linguistic explanations. We will demonstrate that explanation methods can generate image explanations with accurate spatial grounding property, meanwhile, reveal more related inputs (pixels of the image input and words of the linguistic sequence input) that are used as evidence for the model decisions. Also, explanation methods can disentangle the contributions of the image and text inputs and provide more interpretable information than purely image-centered attention.

With explanation methods \cite{SamPIEEE21}, we have a deeper understanding of image captioning models beyond visualizing the attention. We also observe that image captioning models sometimes hallucinate words from the learned sentence correlations without looking at the images and sometimes use irrelevant evidence to make predictions. The hallucination problem is also discussed in \cite{OBJECTHALLU:rohrbach2018object}, where the authors state that it is possibly caused by language priors or visual mis-classification, which could be partially due to the biases present in the dataset. The image captioning models tend to generate those words and sentence patterns that appear more frequently during training. The language priors are helpful, though, in some cases. \cite{SceneGraph:yang2019auto} incorporates the inductive bias of natural language with scene graphs to facilitate image captioning. However, language bias is not always correct, for example, not only men ride snowboards \cite{GenderBias:hendricks2018women} and bananas are not always yellow \cite{RUBI_VQA_bias:cadene2019rubi, HINT:selvaraju2019taking}. 
To this end, \cite{GenderBias:hendricks2018women} and \cite{HINT:selvaraju2019taking} attempted to generate more grounded captions by guiding the model to make the right decisions using the right reasons. They adopted additional annotations, such as the instance segmentation annotation and the human-annotated rank of the relevant image patches, to design new losses for training.

In this paper, we reduce object hallucination by a simple \emph{\textbf{LRP-inference fine-tuning}} (LRP-IFT) strategy, without any additional annotations. We firstly show that the explanations, especially LRP, can weakly differentiate the grounded (true-positive) and hallucinated (false-positive) words. Secondly, based on the findings that LRP reveals the related features of the explained words and that the sign of its relevance scores indicates supporting versus opposing evidence (as shown in Figure~\ref{fig:intro_explanation}), we utilize LRP explanations to design a re-weighting mechanism for the context representation. During fine-tuning, we up-scale the supporting features and down-scale the opposing ones using a weight calculated from LRP relevance scores. Finally, we use the re-weighted context representation to predict the next word for fine-tuning.

LRP-IFT is different from standard fine-tuning which weights the gradients of parameters with small learning rates to gradually adapt the model parameters. Instead, it pinpoints the related features/evidence for a decision and guides the model to tune more on those related features. This fine-tuning strategy resembles how we correct our cognition bias. For example, when we see a green banana, we will update the color feature of bananas and keep the other features such as the shape. 

We will demonstrate that LRP-IFT can help to de-bias image captioning models from frequently occurring object words. Though language bias is intrinsic, we can guide the model to be more precise when generating frequent object words rather than hallucinate them.
We implement the LRP-IFT on top of pre-trained image captioning models trained with Flickr30K \cite{FLICKR30K:young2014image} and MSCOCO2017 \cite{MSCOCO:lin2014microsoft} datasets and effectively improve the mean average precision (mAP) of predicted frequent object words evaluated across the test set.
At the same time, the overall performance in terms of sentence-level evaluation metrics is maintained. 

The contributions of this paper are as follows: 

\begin{itemize}
    \item We establish explanation methods that disentangle the contributions of the image and text inputs and explain image captioning models beyond visualizing attention.
    \item 
    We quantitatively measure and compare the properties of explanation methods and attention mechanisms, including tasks of finding the related features/evidence for model decisions, grounding to image content, and the capability of debugging the models (in terms of providing possible reasons for object hallucination and differentiating hallucinated words).
    \item We propose an LRP-inference fine-tuning strategy that reduces object hallucination and guides the models to be more precise and grounded on image evidence when predicting frequent object words. Our proposed fine-tuning strategy requires no additional annotations and successfully improves the mean average precision of predicted frequent object words.
\end{itemize}

In the rest of this paper, Section \ref{sec:related work} introduces recent image captioning models, the state-of-the-art explanation methods for neural networks, and other related works. In Section \ref{sec:imgcapbackground}, we will introduce the image captioning model structures applied in this paper. The adaptations of explanation methods to attention-guided image captioning models are summarized in Section \ref{sec:methodology}. The analyses of attention and explanations and our proposed LRP-inference fine-tuning strategy are introduced in Section \ref{sec:experiments}.

\section{Related Work}
\label{sec:related work}
\subsection{Image Captioning}

Many models adopt the encoder-decoder approach to bridge the gap between image and text, usually with a CNN as the image encoder and an RNN as the sentence decoder \cite{ SHOWTELL:vinyals2015show,DEEPVS:karpathy2015deep,CNNLSTM:soh2016learning}. Considering that it might be helpful to focus on a sub-region of the image when generating a word of the caption, various attention mechanisms have been developed, guiding the model to focus on the relevant parts of the image when predicting a word. 
Some representative works include hard or soft attention \cite{SHOWATTENDTEL:Lxu2015show}, semantic attention \cite{MSM:you2016image}, adaptive attention \cite{KNOWING:lu2017knowing}, bottom-up and top-down attention \cite{BUTD:Anderson_2018_CVPR}, adaptive attention time \cite{AATAttention:huang2019adaptively}, hierarchical attention \cite{HierarchicalAttention:wang2019hierarchical}, X-Linear attention \cite{XLINERAttention:pan2020x}, and spatio-temporal memory attention \cite{SpatioTemporalAttention:ji2020spatio}.
Recently, many works build the attention mechanism with the multi-head attention originated from Transformer models \cite{TRANSFORMER:vaswani2017attention}, such as attention on attention \cite{AOATransformer:huang2019attention}, entangled transformer \cite{EntangledTransformer:li2019entangled}, multi-modal transformer \cite{MultimodelTransformer:yu2019multimodal}, meshed-memory transformer \cite{MeshedMemoTransformer:cornia2020meshed}. These attention mechanisms effectively facilitate image captioning models to better recognize and locate the objects in an image. We will analyze the adaptive attention mechanism \cite{KNOWING:lu2017knowing, BUTD:Anderson_2018_CVPR,AATAttention:huang2019adaptively, HierarchicalAttention:wang2019hierarchical} and the multi-head attention mechanism \cite{TRANSFORMER:vaswani2017attention,ConceptualCaptionTransformer:sharma2018conceptual,AOATransformer:huang2019attention, EntangledTransformer:li2019entangled, MultimodelTransformer:yu2019multimodal, MeshedMemoTransformer:cornia2020meshed}. Both attention mechanisms are employed as a sub-module in a number of works.

Recognizing and locating the objects in an image is often not sufficient to generate fine-grained captions.
In addition to studying attention mechanisms, a branch of research explores the relations of objects (e.g. \emph{playing with} balls) and object attributes (e.g. a \emph{wooden} desk).
Many of these works build a graph to capture the relation and attribute representations of objects, such as the scene graph \cite{SceneGraph:yang2019auto, Transformobjectsintowords:herdade2019image,ASG:chen2020say, HIP:yao2019hierarchy, yao2018exploring} and visual relation graph \cite{CGVRG:shi2020improving}. Some other works aim to generate more fine-grained captions by learning global and local representations in a distilling fashion \cite{ExploreandDistil:liu2019exploring}, by gradually learning the representation via context-aware visual policy \cite{zha2019context}, by parsing and utilizing the noun chunks in the reference captions \cite{shoecontroltell:cornia2019show}. The unified VLP \cite{UnifyRepresentation:zhou2020unified} learns unified image-text representations in the spirit of the BERT embedding \cite{BERT:devlin2019bert}. VIVO \cite{VIVO_NOC:hu2020vivo} and OSCAR \cite{OSCAR:li2020oscar} further enhance the unified representation by incorporating external image-tag pairs for training. These unified representations can be used in various visual-language tasks. \cite{HumanConsensus:wang2020human} uses additional rank annotations of the referenced captions. 

There are also other challenging directions of image captioning like novel object captioning (NOC) and captioning with different styles. NOC tries to predict novel objects that are not in the image-caption training pairs, which overcomes the limitation of fixed training vocabulary and achieves better generalization \cite{NBT_NOC:lu2018neural,NOC:venugopalan2017captioning, POINGTING_NOC:li2019pointing, DCC_NOC:hendricks2016deep, CascadedNOC:feng2020cascaded, VIVO_NOC:hu2020vivo, NONCAP_NOC:agrawal2019nocaps}. \cite{MSCAP:guo2019mscap} and \cite{ GOODNEWS:biten2019good} attempt to generate captions with controlled sentiments and styles.

\subsection{Towards de-biasing visual-language models}
The intrinsic composition of the training data can lead to biased visual-language models. To this end, many works aim to reduce model bias and improve the grounding property of visual-language models. For visual-question-answering (VQA) models, \cite{RUBI_VQA_bias:cadene2019rubi} learns the language bias in advance by using the textual question-answer pairs in order to increase the loss computation for biased answers during training. \cite{LOOKANSWER_VQA_bias:agrawal2018don} proposes a grounded visual question answering model that disentangles the yes/no questions and visual concept-related questions. Both show an effective reduction of the bias for the VQA models.
As for image captioning models, \cite{GenderBias:hendricks2018women} designs an appearance confusion loss and a confidence loss using segmentation annotations to reduce the gender bias of the captioning models.
\cite{HINT:selvaraju2019taking} adopts external human-annotated attention maps to guide the model to generate more grounded captions. Different from the above methods, we propose an LRP-inference fine-tuning strategy that requires no additional annotations to mitigate the influence of language bias for image captioning models. The guidance comes from the explanation scores obtained from explanation methods.  

\subsection{Explanation methods for image captioning models.} 

Many explanation methods explain the predictions of DNNs such as gradient-based methods \cite{GRADIENT:simonyan2013deep, GuidebackPropagation:springenberg2014striving, IntegratedGradient:sundararajan2017axiomatic, GRADCAM:selvaraju2017grad}, decomposition-based methods \cite{DEEPTAYLOR:Lmontavon2017explaining, LRP:bach2015pixel, LRP-LSTM:arras2017explaining, DEEPLIFT:shrikumar2017learning, PATTERN:kindermans2017learning, SHARP:lundberg2017unified, CD:murdoch2018beyond}, and sampling-based methods \cite{LIME:ribeiro2016should, VISUALDEEPNEURAL:zintgraf2017visualizing,MEANINFPERTUR:fong2017interpretable,RISE:Petsiuk2018RISERI,CEMAF:luss2019generating}. 
These explanation methods have provided plausible explanations for various DNN architectures including CNNs \cite{LRP:bach2015pixel, GRADCAM:selvaraju2017grad, DEEPLIFT:shrikumar2017learning, LIME:ribeiro2016should, PATTERN:kindermans2017learning, SHARP:lundberg2017unified, VISUALDEEPNEURAL:zintgraf2017visualizing, MEANINFPERTUR:fong2017interpretable,DECONVNET:zeiler2010deconvolutional}, RNNs \cite{LRP-LSTM:arras2017explaining, CD:murdoch2018beyond, LIME:ribeiro2016should, SHARP:lundberg2017unified}, graph neural networks (GNNs) \cite{GNNLRP:schnake2020xai, GNNexplainer:ying2019gnnexplainer, ExplainGNN:li2020explain, RelEx:zhang2020relex, GraphLIME:huang2020graphlime},
and clustering models \cite{Explainkmeans:kauffmann2019clustering}, making it practical to derive the explanation methods for image captioning models. However, to our best knowledge, only a few works have studied the interpretability of image captioning models so far. In principle, gradient-based methods can be directly applied to image captioning models. Grad-CAM and Guided Grad-CAM have been used to explain non-attention image captioning models \cite{GRADCAM:selvaraju2017grad}. 
\cite{TDvisualsaliency:ramanishka2017top} introduces an explanation method for video captioning models. They further adapt the method to image captioning models by slicing an image with grids to form a sequence of image patches, treated as video frames, however, the slicing operation may cut through object structures. 
Attention heatmaps are usually considered as explanations of image captioning models. The question to what extent attention is suitable as an explanation has been raised in the natural language processing context \cite{ATTENTION_NOT_EXPLANATION:jain2019attention,ATTENTION_NOT_NOT_EXPLANATION:wiegreffe2019attention, IS_ATTENTION_INTERPRETABLE:serrano-smith-2019-attention}. For the image captioning task, although attention heatmaps can show the locations of object words, they cannot disentangle the contributions of the image and text inputs. Furthermore, attention heatmaps meet difficulties to provide pixel-wise explanations that reflect the positive and negative contributions of pixels and regions. These issues can be addressed by several explanation methods. For the sake of keeping the scope of analyses within reasonable limits, we will adapt exemplarily LRP, Grad-CAM, and Guided Grad-CAM to image captioning models.  

\subsection{Explanation-guided training}
Recently, some studies observe that explainable AI is not limited to providing post-hoc insights into neural networks but can also be applied to train a model. \cite{Grad-Camloss:halliwell2020trustworthy} utilizes the saliency maps of Grad-CAM and Guided Grad-CAM to design a pixel-wise cross-entropy loss for transfer learning. They show that the pixel-wise cross-entropy loss can guide the model to make the right decisions using the right reasons, meanwhile, improve image classification accuracy. \cite{HINT:selvaraju2019taking} also uses Grad-CAM saliency maps together with additional human-annotated attention maps to design a ranking loss for image captioning models. They show that the ranking loss can help to generate more grounded captions and maintain sentence fluency. \cite{Few-shot-explanation:sun2020explanation} adopts LRP explanations to guide few-shot classification models. They demonstrate that explanation-guided training can improve the model generalization and classification accuracy for cross-domain datasets. We will show that LRP explanations can also help to mitigate the influence of language bias for image captioning models.

\section{Backgrounds of Image Captioning Models}
\label{sec:imgcapbackground}

\subsection{Notations for image captioning models}
In this section, we recapitulate common structures of image captioning models, which consist of an image encoder, a sentence decoder, and a word predictor module. 
To caption a given image, we first encode the image with pre-trained CNNs or detection modules such as a Faster RCNN and extract a visual feature $\bm{I} \in \mathbbm{R}^{n_v \times d_v}$, where $n_v$ and $d_v$ are the number and dimension of the visual feature. For $\bm{I}$ from a Faster R-CNN, $n_v$ would be the number of regions of interest (ROIs), and for $\bm{I}$ from a CNN, $n_v$ would be the number of spatial elements in a feature map.
Then, the visual feature $\bm{I}$ is decoded by an LSTM augmented with an attention mechanism to generate a context representation. Finally, the word predictor takes the context representation and the hidden state of the decoder as inputs to predict the next word. 

During training, there is a reference sentence as the ground truth, $\mathcal{S} = (w_t)_{t=1}^{l}$, where $w_t$ is a word token, and $l$ is the sentence length. At each time step $t$, the LSTM updates the hidden state $\bm{h}_t$ and memory cell $\bm{m}_t$ as follows.
\begin{align}
    \bm{x}_t &= [\bm{E_m}(w_{t-1}), \bm{I}_g] \\
    \bm{h}_t, \bm{m}_t & = LSTM(\bm{x}_t, \bm{h}_{t-1}, \bm{m}_{t-1})
\end{align}
where $[\cdot]$ denotes concatenation, $\bm{E_m}$ is a word embedding layer that encodes words to vectors, $\bm{E_m}(w_{t-1}) \in \mathbbm{R}^{d_w}$. $\bm{I}_g = 1/n_v \sum_{k=1}^{n_v} \bm{I}_{(k)}$ represents an averaged global visual feature.
During inference, the $w_{t-1}$ is the predicted word from the last step.
Then, an attention mechanism $ATT(\cdot)$ uses $\bm{h}_t$ and $\bm{I}$ to generate a context representation $\bm{c}_t$ for word prediction.
\begin{align}
    \bm{c}_t & = ATT(\bm{h}_t, \bm{I})\\
    \bm{p}_t & = Predictor(\bm{h}_t, \bm{c}_t)
\end{align}
where $\bm{p}_t$ is the predicted score over the vocabulary.
The concrete implementations of $\bm{E_m}$, $\bm{I}_g$, $ATT(\cdot)$, and $Predictor$ may vary across different models. 

\subsection{Attention mechanisms used in this study}
We choose two representative attention mechanisms, adaptive attention \cite{KNOWING:lu2017knowing} and a modified multi-head attention \cite{TRANSFORMER:vaswani2017attention, AOATransformer:huang2019attention}. They are employed in variants by several image captioning models, thus aiming at generalizability for our studies.

\subsubsection{Adaptive attention mechanism}
The adaptive attention mechanism generates a context representation by calculating a set of weights over the visual feature $\bm{I}$ and a sentinel feature $\bm{s}_t$ that represents the textual information. At time step $t$:
\begin{align}
\bm{s}_t &= \sigma(\bm{W}_x\bm{x}_t + \bm{W}_h\bm{h}_{t-1}) \odot \tanh(\bm{m}_t) \label{equ:st}
\end{align}
$\bm{W}_{x} \in \mathbbm{R}^{d_h \times d_x}$ and $\bm{W}_{h} \in \mathbbm{R}^{d_h \times d_h}$ are trainable parameters. $d_h$ and $d_x$ denote the dimension of the hidden state and $\bm{x}_t$, respectively. $\sigma$ denotes the \emph{sigmoid} function. The weights for $\bm{I}$ and $\bm{s}_t$ are calculated as follows:
\begin{align}
    \bm{a} &= \bm{w}_a \tanh(\bm{I}\bm{W}_I + \bm{W}_g\bm{h}_t)\\
    b & = \bm{w}_a \tanh(\bm{W}_s \bm{s}_t + \bm{W}_g\bm{h}_t))\\
        \bm{\alpha}_t &= \mathrm{softmax}(\bm{a}) \in \mathbbm{R}^{n_v} \label{equ:ada_alpha}\\
    \beta_t & = \mathrm{softmax}([\bm{a},b])_{(n_v+1)} \in \mathbbm{R}^{1} \label{equ:ada_beta}\\
    \bm{c}_t &= (1-\beta_t)\sum\nolimits_{k=1}^{n_v} \alpha_{t_{(k)}} \bm{I}_{(k)}  + \beta_t \bm{s}_t \label{equ:ada_hatc}
\end{align}
where $\bm{W}_{I} \in \mathbbm{R}^{d_h \times n_v}$, $\bm{W}_{s}$ and $\bm{W}_{g} \in \mathbbm{R}^{n_v \times d_h}$, $\bm{w}_{a} \in \mathbbm{R}^{n_v}$ are trainable parameters\footnote{Adaptive attention mechanism encodes the visual feature $\bm{I}$ with the same dimension as the hidden state, $d_v = d_h$, $\bm{I} \in \mathbbm{R}^{n_v \times d_h}$.}. $\bm{\alpha}_t \in \mathbbm{R}^{n_v}$ is the attention weight for $\bm{I}$. It tells the model which regions within the image to use for generating the next word. $\beta_t$ is the $(n_v + 1)^{th}$ element of the \emph{softmax} over $[\bm{a},b]$, corresponding to the weight for the component $b$. It balances the visual and textual information used to predict the next word. We use the following expression to summarize the adaptive attention mechanism.
\begin{equation}
    \bm{c}_t = ATT_{ada}(\bm{h}_t, \bm{s}_t,\bm{I}) \label{equ:adaattention}
\end{equation}

\subsubsection{Multi-head attention mechanism}

The multi-head attention is defined with a triplet of query ($\bm{Q}$), key ($\bm{K}$), and value ($\bm{V}$). 
To apply the multi-head attention to the sentence decoder, we adopt $\bm{h}_t$ as the query and two distinct linear projections of $\bm{I}$ as $\bm{K}$ and $\bm{V}$.
\begin{equation}
    \bm{Q} = \bm{h}_t, \bm{K} = \bm{I}\bm{W}_k, \bm{V} = \bm{I}\bm{W}_v
\end{equation}
where $\bm{W}_K, \bm{W}_V \in \mathbbm{R}^{d_v \times d_h}$. We evenly split the hidden dimension $d_h$ to obtain multiple triplets of ($\bm{Q}^{(i)}$, $\bm{K}^{(i)}$,  $\bm{V}^{(i)}$), denoted as multiple heads. For each head, the attention weight over $\bm{V}^{(i)}$ is the scaled dot product of $\bm{Q}^{(i)}$ and $\bm{K}^{(i)}$ and we can obtain a weighted feature $\bm{v}^{(i)}$ as follows.
\begin{equation}
    \begin{aligned}
    \bm{\alpha}^{(i)}& = \mathrm{softmax}(\frac{\bm{Q}^{(i)}\bm{K}^{(i)\bm{T}}}{\sqrt{d_h/n_h}}) \in \mathbbm{R}^{n_v}\\
    \bm{v}^{(i)} &= \sum_{k=1}^{n_v} \bm{\alpha}^{(i)}_{(k)} \bm{V}^{(i)}_{(k)} \in \mathbbm{R}^{d_h/n_h}
    \end{aligned} \label{equ:mhselfattention}
\end{equation}
where $n_h$ is the number of head\footnote{In most of the works using multi-head attention, $d_h$ is divisible by $n_h$.}. By concatenating the weighted feature of each head, we can obtain the integral attended feature $\bm{v}$, which is further fed to a linear layer to generate the visual representation. 
\begin{equation}
    \begin{aligned}
        \bm{v} &= [\bm{v}^{(0)},\dots,\bm{v}^{(n_h)}] \in  \mathbbm{R}^{d_h}\\
    \bm{\hat{v}} &= \bm{W_v} \bm{v} + \bm{b}_v
    \end{aligned}
\end{equation}
where $\bm{W}_v \in \mathbbm{R}^{d_h \times d_h}$ and $\bm{b}_v \in \mathbbm{R}^{d_h}$ are trainable parameters.

Under the image captioning setup, there are cases where the visual feature is less relevant to the predicted word, e.g. ``a'' and ``the''. Thus, we add another gate to control the visual information, which is consistent with many recent image captioning models using the multi-head attention module \cite{AOATransformer:huang2019attention, EntangledTransformer:li2019entangled, MeshedMemoTransformer:cornia2020meshed}. This also shares the same spirit of $\beta_t$ in the adaptive attention mechanism, 
which controls the proportion of image and textual information. Specifically, we generate the gate using the hidden state and the gated output $\bm{c}_t$ is the context representation for prediction.
\begin{equation}
    \bm{c}_t = \sigma(\bm{W}_{mh} \bm{h}_t + \bm{b}_{mh}) \odot \bm{\hat{v}}
\end{equation}
where $\bm{W}_{mh} \in \mathbbm{R}^{d_h \times d_h}$ and $\bm{b}_{mh} \in \mathbbm{R}^{d_h}$ are trainable parameters and $\sigma$ is the \emph{sigmoid} function. We briefly summarize the multi-head attention mechanism as follows.
\begin{equation}
    \bm{c}_t = ATT_{mha}(\bm{h}_t, \bm{I})
\end{equation}

\begin{figure*}
    \centering
    \includegraphics[width=0.9\textwidth]{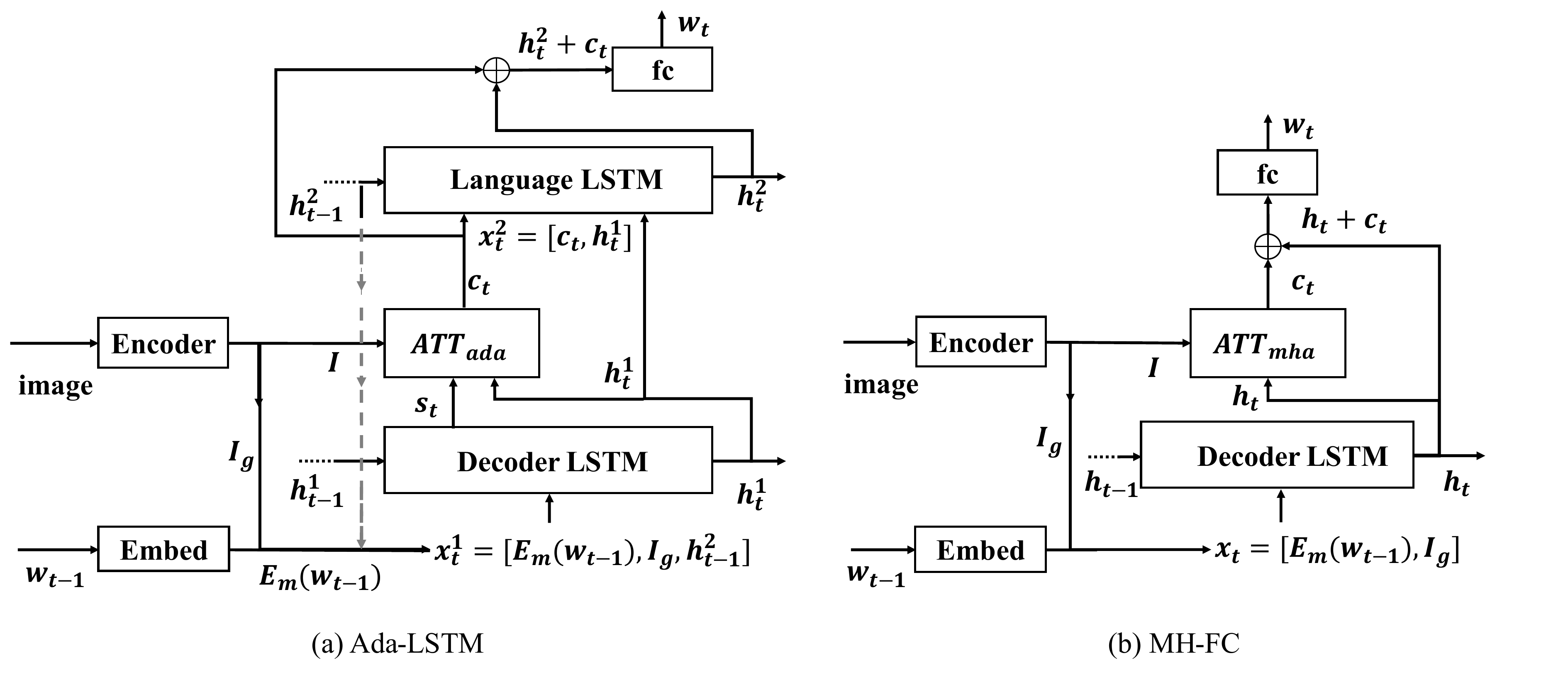}
    \caption{The model structures of two image captioning models. (a): The Ada-LSTM model with the adaptive attention mechanism and an \emph{LSTM + fc} module as the word predictor. (b): the MH-FC model with the multi-head attention mechanism and an \emph{fc} layer as the word predictor.}
    \label{fig:ada_mha}
\end{figure*}
\subsubsection{Image captioning models with adaptive attention and multi-head attention}
We build two image captioning models in this paper. The details of the two models are illustrated in Figure \ref{fig:ada_mha}. The left of Figure \ref{fig:ada_mha} is the \textbf{Ada-LSTM} model that consists of an adaptive attention module and an LSTM followed by a fully connected (\emph{fc}) layer as the word predictor. Note that the $\bm{x}_t$ is adjusted accordingly to incorporate the predictor. On the right is the \textbf{MH-FC} model that adopts a multi-head attention module followed by an \emph{fc} layer as the word predictor. Both model structures are commonly used \cite{KNOWING:lu2017knowing, BUTD:Anderson_2018_CVPR, AOATransformer:huang2019attention,zha2019context,shoecontroltell:cornia2019show}.

The image captioning models are usually trained with cross-entropy loss in the first stage:
\begin{equation}
    \mathcal{L} = \mathcal{L}_{ce}(\bm{p}, \bm{y}) \label{equ:standard_ce}
\end{equation}
where $\bm{p} = (\bm{p}_t)_{t=0}^l$ is the predicted scores over vocabulary, $l$ is the sentence length, and $\bm{y}$ is the ground truth label of a referenced caption. Then, the models are further optimized with the \emph{SCST} algorithm from \cite{SCST:rennie2017self}. SCST optimizes non-differentiable evaluation metrics, e.g. CIDEr score \cite{CIDEr:Vedantam_2015_CVPR}, using reinforcement learning: 
\begin{equation}
    R = \mathbbm{E}_{S^s,S^{greedy} \backsim \bm{p}}[\text{metric}(S^s,S^{gt})-\text{metric}(S^{greedy}, S^{gt})] \label{equ:standard_cider}
\end{equation}
where $R$ is the reward, $S^s$ is the sampled sentence from the predicted distribution $\bm{p} = (\bm{p}_t)_{t=0}^l$, $S^{greedy}$ is the predicted sentence with greedy search, and $S^{gt}$ is the referenced caption. The training objective is to obtain higher reward $R$. CIDEr is usually adopted to calculate the reward and some papers also call this algorithm as CIDEr optimization \cite{BUTD:Anderson_2018_CVPR,AOATransformer:huang2019attention}.

\section{Explanation methods for image captioning models}
\label{sec:methodology}
In this section, we will explain how to adapt LRP \cite{LRP:bach2015pixel}, Grad-CAM, and Guided Grad-CAM \cite{GRADCAM:selvaraju2017grad} for use in attention-guided image captioning models. For brevity, we will use Grad* to denote Grad-CAM and Guided Grad-CAM. 

Grad* methods are based on gradient backpropagation and can be directly applied to the attention-guided image captioning models. Grad* methods first backpropagate the gradient of a prediction till the visual feature $\bm{I}$, denoted as $g(\bm{I}) \in \mathbbm{R}^{n_v \times d_v}$. Then, we can obtain a channel-wise weight from $g(\bm{I})$ for the visual feature $\bm{I}$, which is $\bm{w}_I = \sum_{k=1}^{n_v} g(\bm{I})_{(k)} \in \mathbbm{R}^{d_v}$. $\bm{I}$ is further summed up over the feature dimension, weighted by $\bm{w}_I$, to generate the class activation map, $CAM =ReLU(\sum_{k=1}^{d_v} w_{I_{(k)}} \bm{I}_{(k)}) \in \mathbbm{R}^{n_v}$, which reflects the importance of each pixel in the feature map. Grad-CAM reshapes and up-samples the class activation map to generate the image explanations. To obtain fine-grained and high-resolution explanations, Grad-CAM is fused with GuidedBackpropagation \cite{GuidebackPropagation:springenberg2014striving} by element-wise multiplication. GuidedBackpropagation can be easily implemented in pytorch by writing a custom {\tt torch.autograd.Function} wrapping the stateless {\tt ReLU} layers. This fused method is Guided Grad-CAM. The linguistic explanations of Grad* methods are obtained by summing up the gradients of the word embeddings.  
Next, we will elaborate on LRP for image captioning models.

We briefly introduce the basics of LRP. For an in-depth introduction, we refer to a book chapter like \cite{montavon2019layer}. 
LRP explains neural networks by assigning a \emph{relevance score} to every neuron within the network. The relevance assignment is achieved by backpropagating the relevance score of a target prediction along the network topology until the inputs according to LRP rules. 

Consider the basic component of neural networks as a linear transformation followed by an activation $f(\cdot)$. 
\begin{equation}
    \begin{aligned}
    z_j &= \sum_i w_{ij}y_i + b_j \label{equ:lrp_linear}\\
    \hat{z}_j &= f(z_j)
    \end{aligned}
\end{equation}
where $y_i$ is the input neuron, $z_j$ is the linear output, and $\hat{z}_j$ is the activation output. We use $R(\cdot)$ to denote the relevance score of a neuron.
Suppose $R(\hat{z}_j)$ is known, we would like to distribute $R(\hat{z}_j)$ to all of its input neurons $y_i$, denoted as relevance attribution $R_{i \leftarrow j}$. We refer to two LRP rules for relevance backpropagation that are frequently applied \cite{LRP:bach2015pixel,SebasIJCNN2020:kohlbrenner2019towards,houidi2020use,hagele2020resolving}:
\begin{enumerate}
    \item $\epsilon$-rule
    \begin{equation}
        R_{i \leftarrow j} = R(\hat{z}_j)\frac{y_iw_{ij}}{z_j + \epsilon \odot \mathrm{sign}(z_j)} \label{equ:lrpruleepsilon}
    \end{equation}
    where $\epsilon$ is a small positive number. The stabilizer term $\epsilon \odot \mathrm{sign}(z_j)$ guarantees that the denominator is non-zero.
    \item $\alpha$-rule
    \begin{equation}
        R_{i \leftarrow j} = R(\hat{z}_j)\left((1+\alpha)\frac{(y_iw_{ij})^+}{z_j^+} - \alpha\frac{(y_iw_{ij})^-}{z_j^-}\right) \label{equ:lrprulealphabeta}
    \end{equation}
    where $\alpha \geqslant 0$, $(\cdot)^+ = \max(\cdot,0)$, and $(\cdot)^- = \min(\cdot,0)$. By separating $y_iw_{ij}$ and $z_j$ into positive and negative parts, the $\alpha$-rule ensures a boundedness of relevance terms. The parameter $\alpha$ determines the ratio of focus on positive and negative contribution during relevance backpropagation, from the output $\hat{z}_j$ to all of its inputs $y_i$.
\end{enumerate}

The relevance of neuron $y_i$ is the summation of all its incoming relevance attribution flows.
\begin{equation}
    R(y_i) = \sum_j R_{i \leftarrow j}
\end{equation}

LRP has provided plausible explanations for CNNs \cite{LRP:bach2015pixel}, RNNs such as LSTM \cite{LRP-LSTM:arras2017explaining}, and also GNNs \cite{GNNLRP:schnake2020xai}. These modules are commonly used in image captioning models. To explain image captioning models with LRP, we define next how to apply LRP to the attention mechanisms. 

From Section \ref{sec:imgcapbackground}, we have seen that attention mechanisms involve non-linear interactions of the visual features and the hidden states of the decoder. However, the attention mechanisms mainly serve as weighting operations for features. Thus, we consider an attention mechanism as a linear combination over a set of features with weights such that LRP relevance scores are not backpropagated through the weights. This is consistent with the ``signal-take-all'' redistribution explored in \cite{arras:ACL2019}. In this way, we can directly apply LRP rules to distribute the relevance score of the context representation to the visual features according to the attention weights and bypass the computations within the attention mechanisms.

\begin{figure*}[tb]
    \centering
    \includegraphics[width=0.9\textwidth]{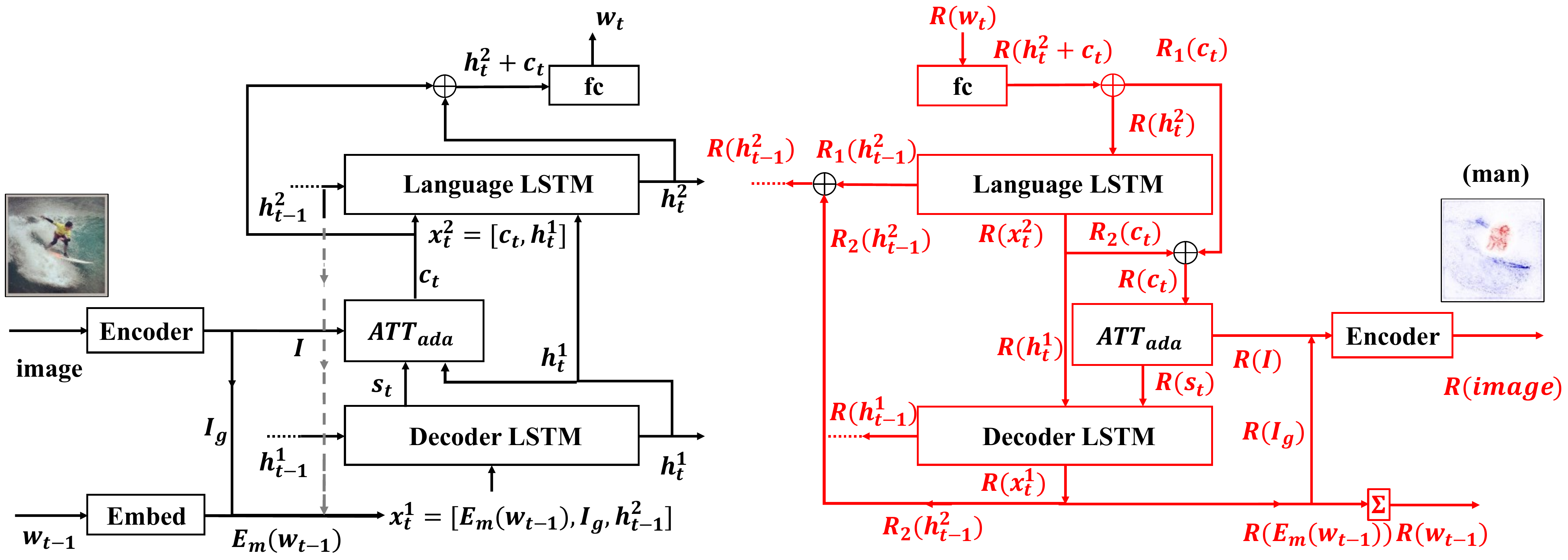}
    \caption{The LRP relevance backpropagation flow path through the Ada-LSTM model.}
    \label{fig:ada-lstm-lrp}
\end{figure*}

To give an overview of LRP for image captioning models, we take the Ada-LSTM model as an example and elaborate on each step of the explanation in Figure \ref{fig:ada-lstm-lrp} and Algorithm \ref{alg: LRP_ada_lstm}. It is important to realize here, that LRP follows topologically the same flow as the gradient backpropagation (except the attention mechanisms) along the edges of a directed acyclic graph. The difference lies in replacing the partial derivatives on the edges by LRP redistribution rules motivated by the deep Taylor framework \cite{DEEPTAYLOR:Lmontavon2017explaining}.

We initialize the relevance score of a target word, $R(w_T)$, from the output of the last \emph{fc} layer (the logits). Then, as illustrated in Figure \ref{fig:ada-lstm-lrp}, LRP-type operations for computing $R(\cdot)$ are applied to the layers \emph{fc}, $\oplus$, \emph{Language LSTM}, $ATT_{ada}$, \emph{Decoder LSTM}, and \emph{Encoder}. The LRP operations used for these layers are shown as the $\Longrightarrow$ in Algorithm \ref{alg: LRP_ada_lstm}. For each word to be explained, LRP assigns a relevance score to every pixel of the input image ($R(image)$) and every word of the sequence input ($R(w_{T-1}),\dots, R(w_1)$). We can visualize the image explanation as a heatmap after averaging $R(image)$ over the channel dimension. The relevance score of each preceding word is the summation of the relevance scores over the word embedding. In the experiments, we will also use the \emph{relevance score} to denote the explanation scores of gradient-based methods.
\begin{algorithm}[tb]
\scriptsize
\caption{LRP for Ada-LSTM model to explain $w_T$. The appearing symbols correspond to those in Figure \ref{fig:ada-lstm-lrp}.
Notations: $\bm{\alpha}_t$ (Eq.~\eqref{equ:ada_alpha}), $\beta_t$ (Eq.~\eqref{equ:ada_beta}), and $\bm{s}_t$ (Eq.~\eqref{equ:st}), $\epsilon$-rule (Eq.~\eqref{equ:lrpruleepsilon}), $\alpha$-rule (Eq.~\eqref{equ:lrprulealphabeta}) , $[\cdot]$ denotes concatenation. 
} 
\label{alg: LRP_ada_lstm}
\begin{algorithmic}[1]
\REQUIRE $R(w_T)$, $\bm{\alpha}_t, \beta_t$
\ENSURE $R(image), R(w_{T-1}), \dots, R(w_1)$
\STATE $R(w_T),\mathrm{fc} \xRightarrow{\epsilon\texttt{-rule}} R(\bm{c}_T + \bm{h}^2_T)$
\STATE $R(\bm{c}_T + \bm{h}^2_T),\oplus \xRightarrow{\epsilon\texttt{-rule}} R_1(\bm{c}_T)$, $R(\bm{h}^2_T)$
\FOR{$t \in [T, \dots, 0, start]$}
    \STATE $R(\bm{h}^2_t),\text{Language-LSTM} \xRightarrow{\epsilon\texttt{-rule}} R_2(\bm{c}_t),R( \bm{h}^1_t), R_1(\bm{h}^2_{t-1})$
    \STATE $R_1(\bm{c}_t)+R_2(\bm{c}_t), ATT_{ada} \xRightarrow{\epsilon\texttt{-rule}} R(\bm{s}_t), R_t(\bm{I})$
    \STATE $R(\bm{h}^1_t), R(\bm{s}_t),\text{Decoder LSTM} \xRightarrow{\epsilon\texttt{-rule}} \underbrace{R(\bm{E_m}(w_{t-1})), R_t(\bm{I}_g), R_2(\bm{h}^2_{t-1})}_{=R(\bm{x}^1_t)},R(\bm{h}^1_{t-1})$
    \STATE $R(\bm{E_m}(w_{t-1})) \xRightarrow{\sum} R(w_{t-1})$
\ENDFOR
\STATE $\sum_t R_t(\bm{I}), \sum_t R_t(\bm{I}_g),\text{CNN} \xRightarrow{\epsilon\texttt{-rule},\alpha\texttt{-rule}} R(image)$
\RETURN $R(image), R(w_{T-1}), \dots, R(w_1)$
\end{algorithmic}
\end{algorithm}

\section{Experiments}
\label{sec:experiments}
\subsection{Model preparation and implementation details}
We train the Ada-LSTM model and the MH-FC model on Flickr30K \cite{FLICKR30K:young2014image} and MSCOCO2017 \cite{MSCOCO:lin2014microsoft} datasets for the following experiments\footnote{\url{https://github.com/SunJiamei/LRP-imagecaptioning-pytorch.git}}.

    \textbf{Dataset}:
    We prepare the Flickr30K dataset as per the Karpathy split \cite{DEEPVS:karpathy2015deep}.
    For MSCOCO2017, we use the original validation set as the offline test set and extract 5000 images from the training set as the validation set. The train/validation/test sets are with 110000/5000/5000 images. Vocabularies are built only on the training set. We encode the words that appear less than 3 and 4 times as an unknown token $<$\textbf{unk}$>$ for Flickr30K and MSCOCO2017, respectively, resulting in 9585 and 11026 vocabularies for the two datasets.

    \textbf{Encoder}: We experiment with CNN and FasterRCNN as the image encoder. The CNN features are extracted from the pre-trained VGG16 \cite{VGG16:Simonyan15} on ImageNet, specifically, we use the output of ``\emph{block5\_conv3}'' with a shape of $14\times14\times512$. The Faster RCNN encoder provides bottom-up image features corresponding to the candidate regions for object detection. We refer to \textbf{Detectron2} \cite{Detectron2:wu2019detectron2}\footnote{\url{https://github.com/airsplay/py-bottom-up-attention.git}} to extract $n_v=36$ features per image with 2048 channels each. Both the CNN features and the bottom-up features are further processed by a linear layer to generate the visual feature $\bm{I} \in \mathbbm{R}^{n_v \times d_v}$.
    
    \textbf{Decoder and predictor}: We train the Ada-LSTM and MH-FC models in Figure \ref{fig:ada_mha}, with $d_v, d_h = 512$ for CNN features and $d_v, d_h = 1024$ for bottom-up features. The word embedding dimension is 512 for all the models. The number of multiple heads for MH-FC model is 8.
    
    \textbf{LRP parameters}: 
    We follow the suggestions of \cite{SebasIJCNN2020:kohlbrenner2019towards} on the best practice for LRP rules. We use $\alpha$-rule for convolutional layers with $\alpha = 0$ and $\epsilon$-rule for fully connected layers and LSTM layers with $\epsilon=0.01$.
    
    \textbf{Training details}: We adopt the Adam optimizer for training, with $\beta_1=0.8$, $\beta_2=0.999$, and a learning rate $lr=0.0005$. We anneal \emph{lr} by 20\% when the CIDEr score does not improve for the last 3 epochs and stop the training when the CIDEr score does not improve for 6 epochs. We further optimize the models with the SCST optimization \cite{SCST:rennie2017self} using CIDEr score with $lr=0.0001$. For the models using CNN features, we also fine-tune the CNN encoder with $lr=0.0001$ before applying the SCST optimization.

Table \ref{tab:model_performance} lists the performance of the Ada-LSTM model and the MH-FC model. We generate the captions with beam search (beam size=3) and report five evaluation metrics of image captioning task: METEOR \cite{METEOR:banerjee2005meteor}, ROUGE-L \cite{ROUGE:lin2004rouge}, SPICE \cite{SPICE:anderson2016spice}, CIDEr \cite{CIDEr:Vedantam_2015_CVPR}, and the $F_{BERT}$(idf) metric of BERTScore \cite{BERT_SCORE:bert-score}. To validate our models, we include the performance of some benchmark image captioning models with similar model structures. \textbf{AdaATT} \cite{KNOWING:lu2017knowing} is the first paper that proposes the adaptive attention mechanism. \textbf{SCST} \cite{SCST:rennie2017self} adapts reinforcement learning to image captioning and optimizes non-differentiable evaluation metrics. \textbf{BUTD} \cite{BUTD:Anderson_2018_CVPR} adopts the bottom-up features and uses an LSTM as the word predictor. We can see that our models are properly trained and achieve comparable performance.

\begin{table*}[tb]
    \centering
    \scriptsize
    \caption{The performance of the Ada-LSTM model and the MH-FC model on the test set of Flickr30K and MSCOCO2017 datasets. The performance of AdaATT, SCST, BUTD models are from the corresponding papers. \emph{BU} and \emph{CNN} denote bottom-up features and CNN features, respectively.
    }
    \begin{tabular}{l r r r r r r r}
        
        \hline
         Flickr30K    & $F_{BERT}$ & CIDEr  & SPICE   & ROUGE-L & METEOR \\ \hline
         Ada-LSTM-CNN &90.56       & 51.54  & 13.87   & 46.79   & 20.18     \\ 
         Ada-LSTM-BU  &90.04       & 63.03  & 16.52   & 49.32   & 21.94     \\ 
         MH-FC-CNN   &90.54       & 53.65  & 14.85   & 46.92   & 20.71   \\ 
         MH-FC-BU    &90.14       & 63.22  & 16.90   & 49.22   & 22.37   \\ 
         AdaATT \cite{KNOWING:lu2017knowing} \emph{(CNN+fc)}
                   &  --            & 53.10   & 14.50    & 46.70    & 20.40 \\\hline
        \hline
         MSCOCO2017&  $F_{BERT}$   & CIDEr  & SPICE   & ROUGE-L & METEOR \\ \hline
         Ada-LSTM-CNN &91.83       & 107.03 & 19.49   & 54.34   & 26.10     \\ 
         Ada-LSTM-BU  &91.01       & 111.87 & 19.17   & 55.04   & 25.93    \\ 
         MH-FC-CNN   &91.85       & 108.16 & 20.10   & 54.42   & 26.45   \\ 
         MH-FC-BU    &91.29       & 120.31 & 21.80   & 56.52   & 28.02   \\ \hline
          \hline
         MSCOCO2014&  $F_{BERT}$   & CIDEr  & SPICE   & ROUGE-L & METEOR \\ \hline
         AdaATT \cite{KNOWING:lu2017knowing} \emph{(CNN+fc)}
                   &  --            & 108.50  & 19.40    & 54.90    & 26.60 \\ 
         SCST:att2all \cite{SCST:rennie2017self} \emph{(CNN+fc)}
                   &  --            & 114.00  & --       & 55.70    & 26.70 \\
         BUTD \cite{BUTD:Anderson_2018_CVPR} \emph{(BU + LSTM)}
                   &  --            & 120.10  & 21.40    & 56.90    & 27.70    \\\hline
    \end{tabular}

    \label{tab:model_performance}
\end{table*}

\begin{figure}
    \centering
    \includegraphics[width=0.5\textwidth]{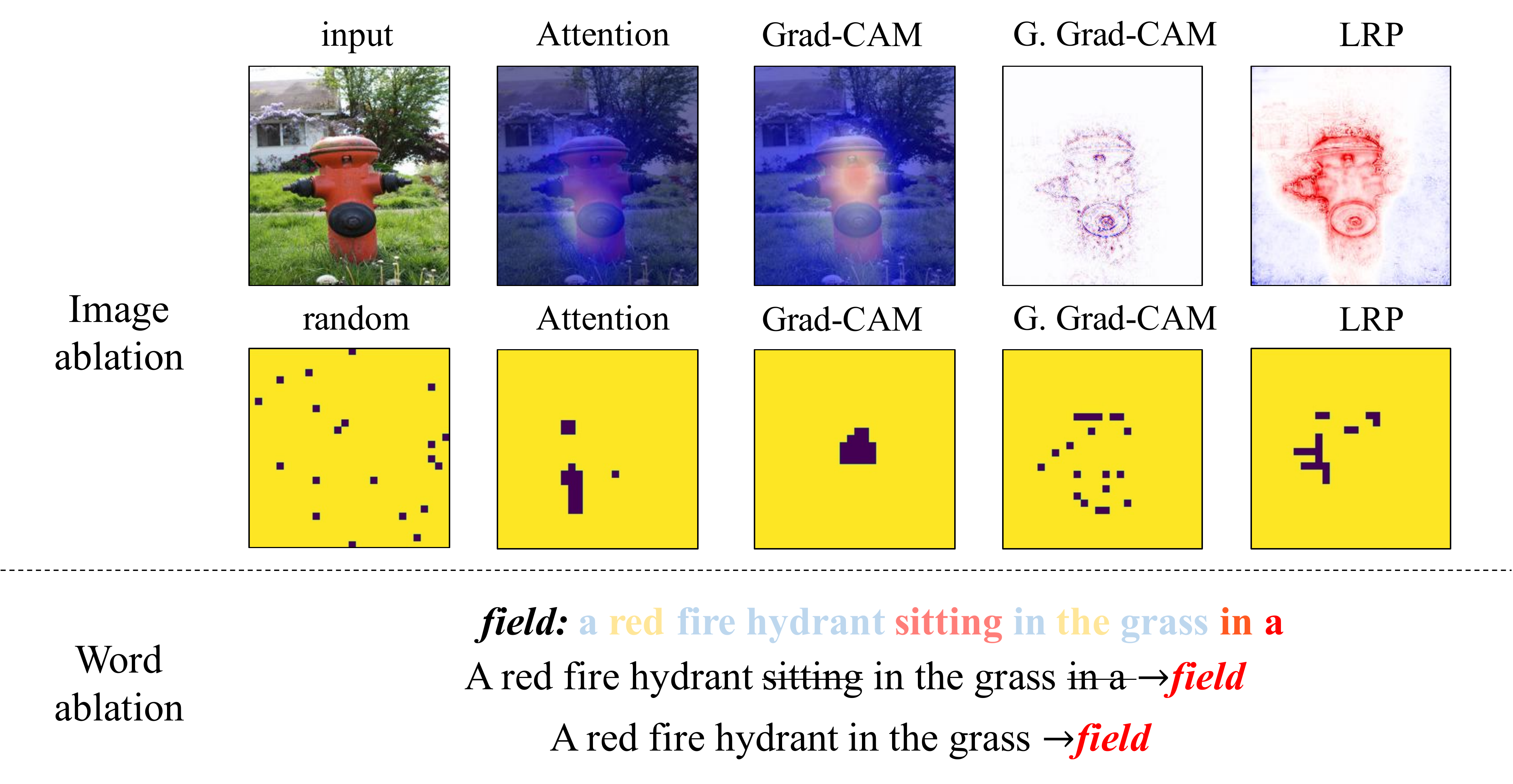}
    \caption{The image ablation (upper) and word ablation (lower) experiment. For image ablation, the first row shows the image explanations of the word \emph{hydrant}, the second row shows the masked patches with high relevance scores.}
    \label{fig:abalation_experiment}
\end{figure}

\subsection{Explanation results and evaluation}
\label{sec:explanation results and evaluation}
Section \ref{sec:introduction} has shown some examples of the explanation results generated by LRP, Grad-CAM, Guided Grad-CAM. In comparison to attention heatmaps, we observe the following. 

Firstly, explanation methods can disentangle the contributions of the image input and the textual input,
which is beyond the interpretability that attention mechanisms can provide. 
Secondly, some explanation methods provide high-resolution, pixel-wise image explanations, such as LRP and Guided Grad-CAM. Thirdly, LRP explicitly shows the positive and negative evidence used by the model to make decisions. 
In the following experiments, we will quantitatively evaluate the information content of attention, LRP, Grad-CAM, and Guided Grad-CAM with two ablation experiments and one object localization experiment. The ablation experiment aims to measure the information in the visual domain and the text domain, expressed by the relevance scores assigned to pixels and words. The object localization experiment evaluates the visual grounding property of relevance scores for image regions. 

\subsubsection{Ablation experiment}
\label{sec:ablationexperiment}
We conduct the ablation experiment for both the image explanations and the linguistic explanations, as illustrated in Figure \ref{fig:abalation_experiment}. We demonstrate the approach using the same example in Section \ref{sec:introduction} based on the caption: \emph{A red fire hydrant sitting in the grass in a field}.

The first row of Figure \ref{fig:abalation_experiment} shows the image explanations of the word \emph{hydrant}, which highlight parts of the image related to the \emph{hydrant}. To assess whether the highlighted areas contribute to the prediction, we firstly segment the image into non-overlapping $8\times8$ patches. Secondly, we sum the relevance scores within each patch as the patch relevance. Thirdly, we mask the top-20 high-relevance patches with the training data mean, to eliminate the contributions of these patches. The top-20 high-relevance patches found by different explanation methods are shown in the second row of Figure \ref{fig:abalation_experiment}. Finally, we predict a caption on the masked image. If the masked areas are important to the prediction, the model will be less confident to predict the target word or will not generate the target word at all from the masked image.

The linguistic explanations reflect the contributions of the previously generated sequence. For example, when generating the word \emph{field}, the model perhaps uses the words ``''\emph{sitting}, ``\emph{in}'', and ``\emph{a}'' as related evidence. Similar to the idea of the image ablation experiment, we remove the top-3 relevant words in the preceding sequence and forward the modified sequence to the model in a teacher-forcing manner. Finally, we observe the new probability of the target word. We do not modify the image for the word ablation experiment. If the removed words are strongly related to the prediction, the new probability of the target word will drop considerably compared to its original value. 

We conduct the ablation experiment using image captioning models trained on the MSCOCO2017 dataset and CNN features. We report the results on the test set. For the word ablation experiment, we consider the predicted words with a sequence index greater than 6 so that there is a sufficiently long preceding word sequence to avoid evaluating purely frequency-based predictions in the experiment. For the image ablation experiment, we consider all the predicted object words. A random ablation is included as a baseline. 

\begin{figure}
    \centering
    \includegraphics[width = 0.5\textwidth]{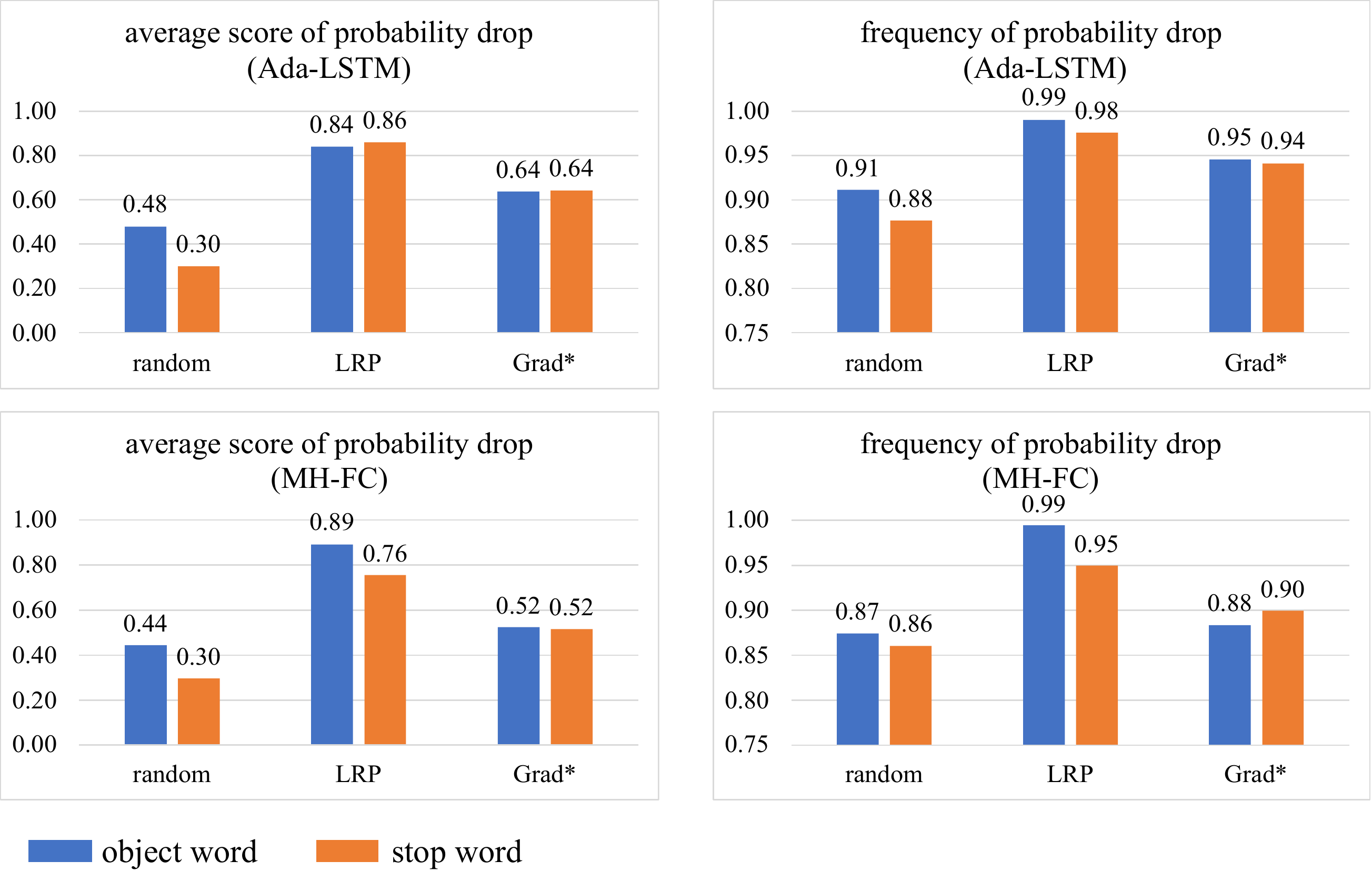}
    \caption{The results of the word ablation experiment on MSCOCO2017 test set. The numbers of evaluated object words and stop-words are 3,710 and 11,686 for the Ada-LSTM model, and 3,359 and 11,512 for the MH-FC model. Higher average scores and higher frequencies are better. 
    }
    \label{fig:word-ablation}
\end{figure}

Figure \ref{fig:word-ablation} shows the results of word-ablation experiments. The words we explain are split into object words and stop-words. We show the frequency of probability drop,
and the difference between the original word probability and the new word probability after the word deletion (denoted as an average score of probability drop).
A higher average score of probability drop means the model is less confident to make the original prediction after ablation, therefore, the ablated words are more strongly related to the prediction. 
LRP and gradient-based explanation methods achieve a decrease in prediction probability more often and with greater impact than the random ablation, indicating that the words found by explanation methods are used by the model as important evidence to predict the target word. LRP achieves both the highest frequency and the highest average score of probability drop.

\begin{figure*}
    \centering
    \includegraphics[width=0.8 \textwidth]{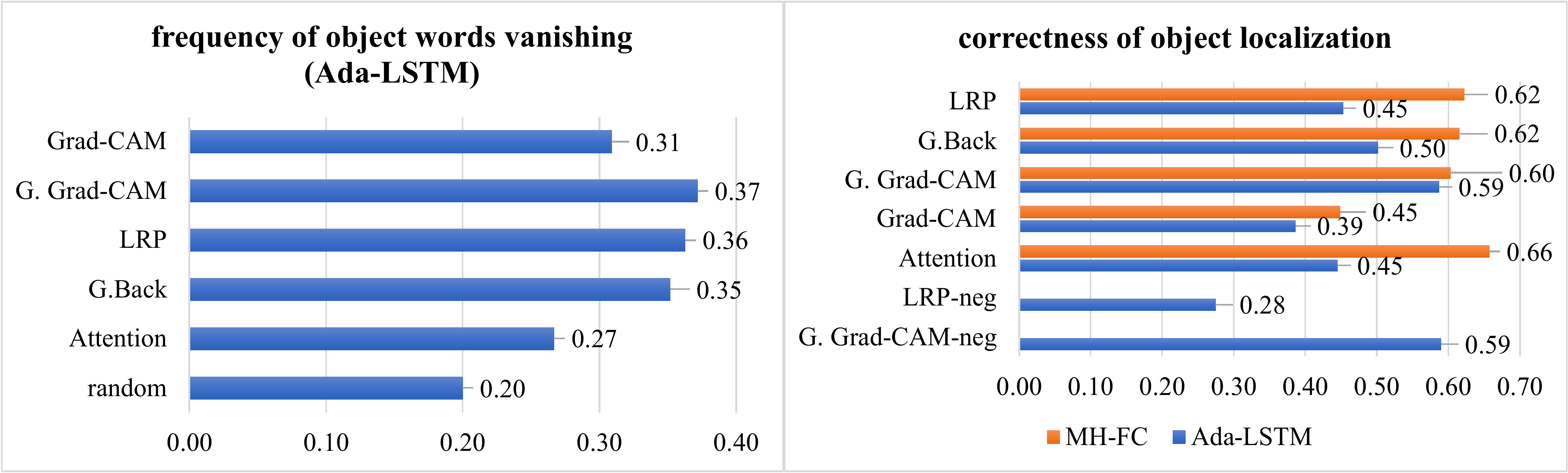}
    \caption{Left: the results of the image ablation experiment using the Ada-LSTM model. There are 9,645 evaluated object words. Higher frequency is better. Right: the average \emph{correctness} of object localization. There are 4,691/4,649 correctly predicted words for the Ada-LSTM/MH-FC model using the MSCOCO2017 test set. Higher \emph{correctness} scores mean better localization. G. denotes Guided. G.Back denotes GuidedBackpropagation.}
    \label{fig:image-ablation-object-location}
\end{figure*}

In our word ablation experiment, we use 8 heads for the multi-head attention mechanism of the MH-FC model, resulting in 8 sets of attention weights. This is computationally too heavy
for use in the image ablation experiment. We, therefore, implement the image ablation experiment with the Ada-LSTM model and show how often the model \emph{fails to} generate the target word after the image ablation, as shown in Figure \ref{fig:image-ablation-object-location} (left). We can see that high-resolution explanations from the evaluated explanation methods LRP, Guided Grad-CAM, and GuidedBackpropagation achieve a higher frequency of object words vanishing, indicating that the highlighted areas are related to the evidence for model decisions. 


With the above experiment results, we verify that using explanation methods adds information compared to relying on attention heatmaps alone.

\subsubsection{Measuring the correlation of explanations to object locations}
\label{sec:localization}
Many studies employ attention heatmaps as a tool to verify the visual grounding property qualitatively  \cite{SHOWATTENDTEL:Lxu2015show, KNOWING:lu2017knowing, AOATransformer:huang2019attention, HierarchicalAttention:wang2019hierarchical, SpatioTemporalAttention:ji2020spatio}. 
In this part, we will quantify the correlation of explanation results to object locations and show that high-resolution explanations can also achieve a high correlation to the object locations. 

To assess the correlation of explanations to object locations, we utilize the bounding box annotations of the MSCOCO2017 dataset and extend the \emph{correctness} measure from \cite{ATTENTIONCORRECTNESS:liu2017attention}, which evaluates the grounding property of attention heatmaps, to the explanation results. For a correctly predicted object word, we first obtain the relevance scores of the image input, $R(image)$, with explanation methods and average $R(image)$ over the channel dimension, resulting in a spatial explanation $\bm{E} \in \mathbbm{R}^{h \times w}$, where $h$ and $w$ are the height and width of the image. We keep the positive scores of $\bm{E}$ for object localization. The $correctness$ is the proportion of the relevance scores within the bounding box.
\begin{align}
    \bm{E}_p &= norm(\max (\bm{E}, 0)) \\
    correctness &= \frac{\sum_{ij \in bbox}\bm{E}_p[i,j] }{\sum_{ij}\bm{E}_p[i,j]} \in [0,1]
\end{align}
where the $norm(\cdot)$ is the normalization with the maximal absolute value. For the MH-FC model with the multi-head attention mechanism, we generate the explanations for each head, $R(image)^{(i)}$, by only backpropagating the relevance scores or gradients through head $i$. The \emph{correctness} of the MH-FC model is the maximum across the $correctness^{(i)}$ of all the heads, i.e.
\begin{equation}
    correctness_{MH-FC} = \max\limits_i(correctness^{(i)})~
\end{equation}
Higher \emph{correctness} means the relevance scores concentrate more within the bounding box, indicating a better grounding property.
Figure \ref{fig:image-ablation-object-location} (right) shows the average $correctness$ of all the correctly predicted object words across the MSCOCO2017 test set, evaluated with image captioning models trained using CNN features.

First of all, the MH-FC model achieves consistently higher \emph{correctness} than the Ada-LSTM model, indicating that there is at least one head of the MH-FC model that accurately locates the object, especially for attention and LRP where there is a large discrepancy of the \emph{correctness} between the Ada-LSTM and the MH-FC models.

Secondly, high-resolution explanations provided by LRP, Guided Grad-CAM, and GuidedBackpropagation achieve comparable or higher \emph{correctness} than attention. 
The notable exception is due to the spatial localization property of the multiple heads in the MH-FC model.
Combining the results of the ablation experiments, explanation methods tend to find parts of objects which correlate well to the prediction.  

Thirdly, to further get insights into the role of the sign of the relevance scores, we calculate the \emph{correctness} using the absolute value of the negative relevance scores, $\bm{E}_n=norm(\max(-\bm{E}, 0))$. As shown in Figure \ref{fig:image-ablation-object-location} (right), the low \emph{correctness} of ``LRP-neg'' and the high $correctness$ of ``G. Grad-CAM-neg'' verifies that the positive/negative sign of LRP relevance scores reveals the support/opposition of a pixel to the predictions, while for Guided Grad-CAM, \emph{both} positive and negative relevance scores are related to the predictions and irrelevant pixels have low absolute relevance scores. 

Last but not least, our \emph{correctness} evaluation results over various explanation methods under the image captioning scenario are consistent with some prior works. GuidedBackpropagation and LRP generate more coherent explanations for MRI data than other gradient-based methods \cite{eitel_miccai2019}, despite failing certain sanity checks postulated in \cite{beenkim_saliency_NIPS2018_8160}. This underlines the importance of considering multiple criteria in contrast to decisions based on selected axiomatic requirements. 
Furthermore, the sign of LRP relevance scores is meaningful \cite{SebasIJCNN2020:kohlbrenner2019towards}. Both properties can be helpful for \emph{model debugging} \cite{beenkim_saliency_NIPS2018_8160, ASSESS:Lapuschkin2019,anders2019analyzing}. In the next section, we will show how we use LRP to \emph{``debug''} and improve image captioning models.

\subsection{Reducing object hallucination with explanation}
\label{sec:reducing hallucinate}
\begin{figure}
    \centering
    \subfloat[Top-20 frequent object words]{\includegraphics[width=0.4\textwidth]{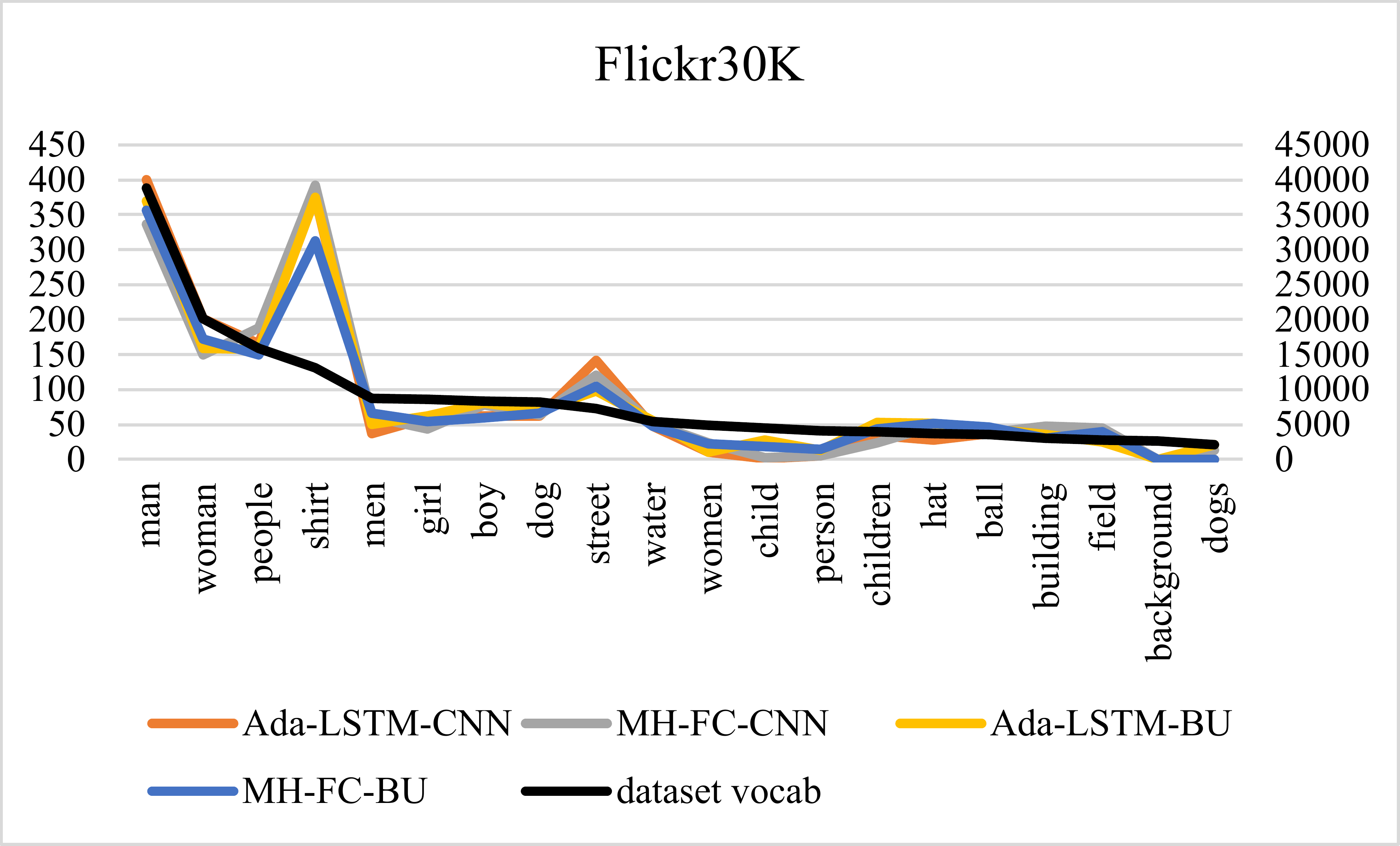}}\\
    \subfloat[Top-25 frequent object words]{\includegraphics[width=0.4\textwidth]{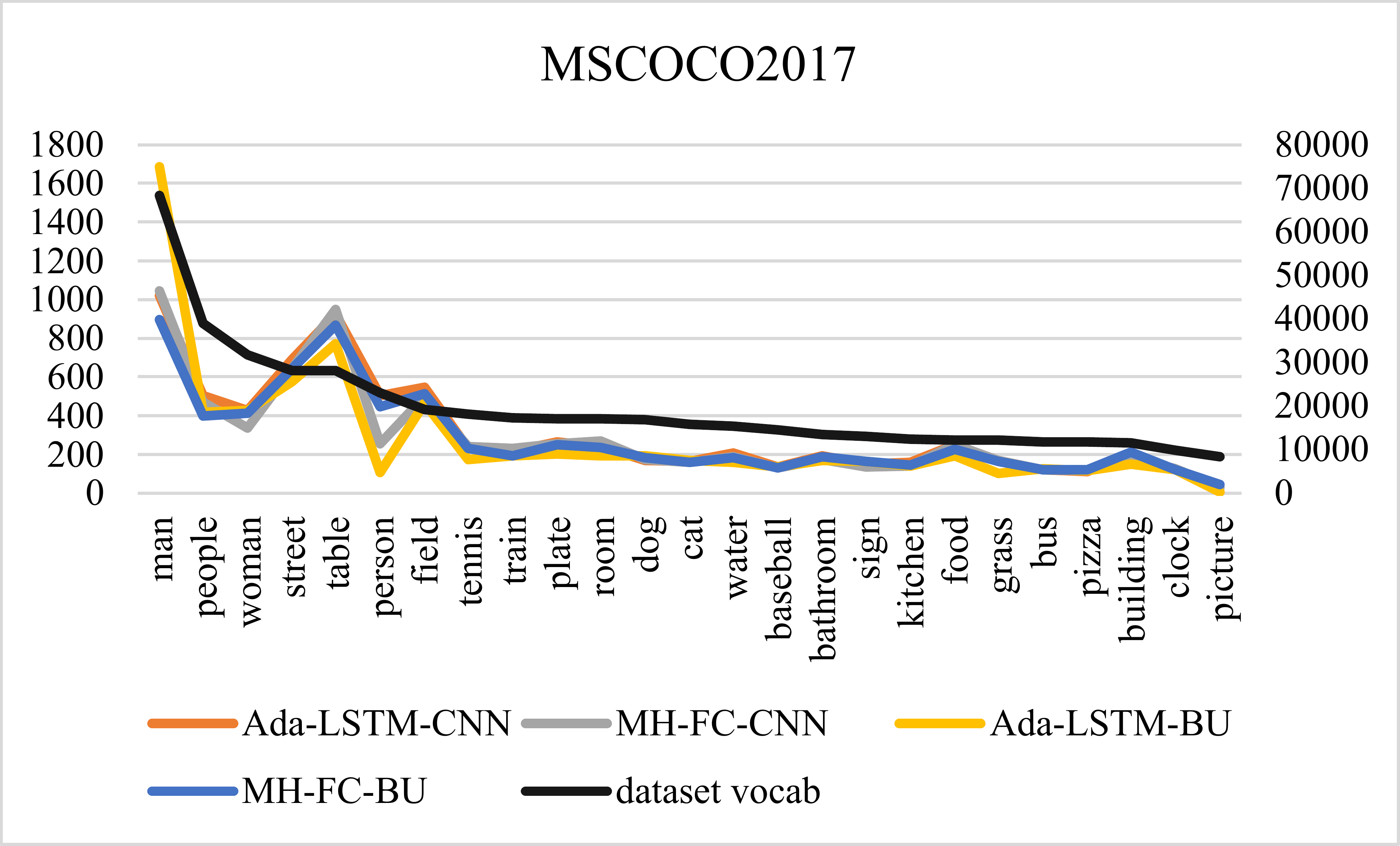}}
    \caption{The counts of top-k frequently appearing object words in Flick30K and MSCOCO2017 training set (right ordinate) and the counts of the predicted object words in the test set (left ordinate).}
    \label{fig:frequentobjectwords}
\end{figure}

In our experiment, we observe the common hallucination problem of image captioning models.
Image captioning models sometimes generate object words that are not related to the image content, which is possibly caused by the learned language priors. The vocabulary and sentence patterns of the \mbox{image-caption} pairs are intrinsically biased toward frequent occurrences. As illustrated in Figure \ref{fig:frequentobjectwords}, the vocabulary count distribution of the predicted words is close to that of the training vocabulary.

A language bias can be helpful for image captioning models. \cite{SceneGraph:yang2019auto} learns the inductive language bias to guide the model to deduce the object relations and attributions. However, it can also cause mistakes. For example, the models could be flawed when predicting gender \cite{GenderBias:hendricks2018women} or always paint \emph{bananas} \emph{yellow}  irrespective of their actual color \cite{RUBI_VQA_bias:cadene2019rubi, HINT:selvaraju2019taking}. 
To this end, we explore the explanations of hallucinated words and investigate using approaches from explainability to reduce object hallucination. 
\begin{figure*}
    \centering
    \includegraphics[width=0.9\textwidth]{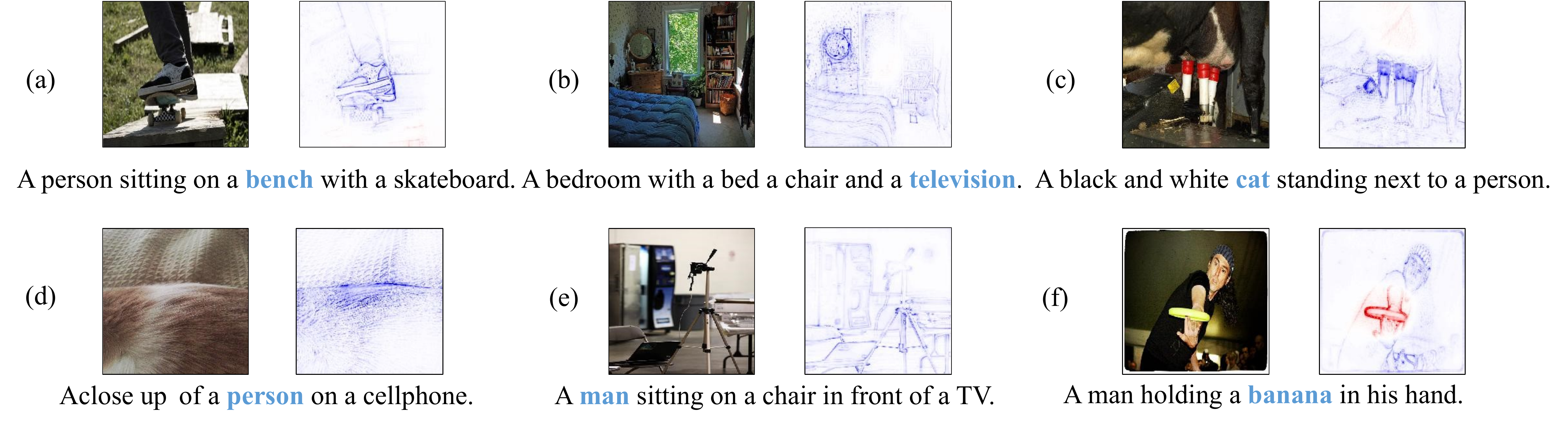}
    \caption{The LRP image explanations of hallucinated words (blue) in the generated image captions. Blue and red pixels indicate negative and positive relevance scores, respectively.}
    \label{fig:hallucinatedwords}
\end{figure*}
\subsubsection{Exploring the explanations of hallucinated words }
\label{sec:rocauc}
Based on the findings in Section \ref{sec:explanation results and evaluation} that high-resolution explanations obtained by LRP and Guided Grad-CAM correlate to the object locations and reflect well the related evidence for predictions, we explore the difference of image explanations between grounded (true-positive) and hallucinated (false-positive) object words. Figure \ref{fig:hallucinatedwords} illustrates some examples of LRP image explanations for hallucinated words.

In Figure \ref{fig:hallucinatedwords} (a) to (e), the LRP image explanations show more negative scores, implying that the model generates hallucinated words mainly with the linguistic information rather than the image information. In Figure \ref{fig:hallucinatedwords} (f), the model mistakes the yellow frisbee for a banana, evidenced by red pixels (positive scores). 

We now quantify the difference in image explanations between true-positive and false-positive object words. Specifically, we use the statistics of image explanations (the $\bm{E}$ mentioned in Section \ref{sec:localization}) to differentiate the hallucinated words. 

We assign a label 1/0 to the true-positive/false-positive predicted words, respectively.
Each word is also assigned with a statistic calculated from the image explanation $\bm{E}$, such as the maximum value (\emph{max($\bm{E}$)}), the 5\% and 50\% quantiles (\emph{quantile-5\%/50\%($\bm{E}$)}), and the mean (\emph{mean($\bm{E}$)}). 
We also evaluate $1-\beta$ from Eq.~\eqref{equ:ada_hatc} of the adaptive attention mechanism. We remind that the adaptive attention mechanism contains a sentinel feature $\bm{s}_t$ that represents the text-dominant information. It then learns a weight, $\beta_t$, which controls the proportion of linguistic information used for predictions. Thus, it is a model-intrinsic baseline to show differences between grounded and hallucinated object words.

\begin{table}[tb]
    \centering
        \scriptsize
    \caption{The AUC scores calculated with different statistics and explanation methods. G. denotes Guided and G.Back denotes GuidedBackpropagation. Higher AUC means the statistic can better differentiate the hallucinated words. The AUC score calculated with $1-\beta$ is \textbf{0.6005}.
    }
    \begin{tabular}{c c c c c c  }
    \hline
         AUC                               & LRP             & G.Grad-CAM & Grad-CAM& Attention & G.Back\\ \hline
        \emph{quantile-5\%}($\bm{E}$)   & \textbf{0.6022}   &  0.4392      & 0.5936    & 0.5598      & 0.4621   \\
        \emph{quantile-50\%}($\bm{E}$)   & 0.5821           &  0.5358      & 0.5730    & 0.5136      & 0.5168   \\
        \emph{max}($\bm{E}$)             & 0.5168           &  0.5743      & 0.5580    & 0.5169      & 0.5575  \\
        \emph{mean}($\bm{E}$)            & 0.5798           &  0.4319      & 0.5857    & 0.5308      & 0.4648  \\
          \hline
    \end{tabular}
    \label{tab:auc}
\end{table}
We calculate the AUC scores, using the labels and statistics of true-positive and false-positive words. A higher AUC score indicates a better differentiation between hallucinated and grounded words. 
Table \ref{tab:auc} lists the AUC scores computed with various explanation methods. We conduct the experiment with the Ada-LSTM model trained on Flickr30K dataset, because its  vocabularies are more imbalanced than that of the MSCOCO2017 dataset. The results are reported on the test set of Flickr30K.
The evaluated words are the top-20 frequent object words\footnote{These most frequent object words are: dogs, building, person, background, field, women, hat, ball, children, child, water, street, boy, dog, girl, men, shirt, people, woman, man.} with 715 false-positive and 1,027 true-positive cases.

The LRP \emph{quantile-5\%($\bm{E}$)} achieves a slightly higher AUC score than $1-\beta$ and can weakly recognize the hallucinated words, which indicates that true-positive words are usually with higher LRP \emph{quantile-5\%($\bm{E}$)} and false-positive words are with lower LRP \emph{quantile-5\%($\bm{E}$)}. The statistics of LRP all obtain AUC scores greater than 0.5, which verifies that the LRP image explanations consist of lower relevance scores for false-positive words, and thus, reflect less supporting evidence for the hallucinated words.

In the next section, we will introduce a fine-tuning strategy that builds upon LRP-based explanations to reduce object hallucination.

\subsubsection{Using LRP explanations to reduce object hallucination}
\label{sec:debias}
We introduce an LRP-inference fine-tuning (LRP-IFT) strategy that can help to de-bias a pre-trained image captioning model and reduce object hallucination. We design a re-weighting mechanism inspired by two properties of LRP explanations: 1) meaningfulness of the positive and negative sign of LRP relevance scores, indicating the support and opposition to the predictions; 2) the property of finding the regions and evidence in the image used by the model to make predictions. 
In particular, we design weights for the input features of the last \emph{fc} layer using the LRP relevance scores and embed the re-weighted features into the model for fine-tuning. 
We elaborate on each step of the fine-tuning strategy with Algorithm \ref{alg: LRP_inference_train} and detail the underlying idea as follows.

To fine-tune an image captioning model $\mathcal{M}$, we generate an initial caption first.
\begin{align}
    (\bm{p}_t)_{t=0}^l & = \mathcal{M}(\mathcal{I})\\
    h(w_t) & = \text{argmax}(\bm{p}_t)
\end{align}
where $\mathcal{I}$ is the image , $\bm{p}_t \in \mathbbm{R}^{V}$ is the probability distribution over the vocabulary at time step $t$, $V$ is the vocabulary size, and $h(w_t)$ is the label of the word $w_t$.

If $w_t$ is \textbf{\emph{not}} a stop-word, we will explain the predicted label $h(w_t)$ through the last \emph{fc} layer using LRP and obtain the relevance scores of the context representation and the hidden state, $R(\bm{c}_t)$ and $R(\bm{h}_t)$. (Remember that $\bm{c}_t + \bm{h}_t$ is the input of the last \emph{fc} layer.)

We then normalize $R(\bm{c}_t)$ and $R(\bm{h}_t)$ with the maximal absolute value, so that their values are in $[-1,+1]$, and generate 
a new word probability distribution $\hat{\bm{p}}_t$ as follows.
\begin{align}
    \bm{\omega}_{\bm{c}_t} &= norm(R(\bm{c}_t)) + 1 \in [0,2] \label{equ:omega_ct}\\
    \bm{\omega}_{\bm{h}_t} &= norm(R(\bm{h}_t)) + 1 \in [0,2] \label{equ:omega_ht}\\
    \hat{\bm{p}}_t &= fc(\bm{\omega}_{\bm{c}_t} \odot \bm{c}_t + \bm{\omega}_{\bm{h}_t} \odot \bm{h}_t) \label{equ:lrp-reweight}
\end{align}

In LRP explanations, positive relevance is attributed to features supporting the prediction of the target class and negative relevance is attributed to contradicting features. The operations performed in Eqs.~\eqref{equ:omega_ct} and~\eqref{equ:omega_ht} construct a weight $\omega$ such that $\omega < 1$ for the opposing features and $\omega > 1$ for the supporting features. The re-weighting mechanism will thus up-scale the supporting features and down-scale the opposing ones. 

\begin{algorithm}[tb]
\scriptsize
\caption{LRP-inference fine-tuning} 
\label{alg: LRP_inference_train}
\begin{algorithmic}[1]
\REQUIRE \text{predicted sequence:}$(w_t)_{t=0}^{l}$, \text{predicted probability:}$(\bm{p}_t)_{t=0}^{l}$
\ENSURE \text{LRP-inference prediction:}$(\bm{\hat{p}}_t)_{t=0}^{l}$
\FOR{$t \in [0, \dots, l]$}
    \IF {$w_t$ \text{ not in stop-words}}
    \STATE  $\bm{p}_t^{h(w_t)}, fc \xRightarrow{\texttt{LRP}} R(\bm{c}_t), R(\bm{h}_t)$
    \STATE  $R(\bm{c}_t) \xRightarrow{\texttt{Eq.~\eqref{equ:omega_ct}}} \bm{\omega}_{\bm{c}_t}$
    \STATE  $R(\bm{h}_t) \xRightarrow{\texttt{Eq.~\eqref{equ:omega_ht}}} \bm{\omega}_{\bm{h}_t}$
    \STATE  $\hat{\bm{p}}_t = fc(\bm{\omega}_{\bm{c}_t} \odot \bm{c}_t + \bm{\omega}_{\bm{h}_t} \odot \bm{h}_t)$
    \ELSE
    \STATE  $\hat{\bm{p}}_t = \bm{p}_t$
    \ENDIF
\ENDFOR
\RETURN $(\bm{\hat{p}}_t)_{t=0}^{l}$
\end{algorithmic}
\end{algorithm}

During fine-tuning, we use the LRP-inference prediction \mbox{$\bm{\hat{p}}=(\bm{\hat{p}}_t)_{t=0}^l$} to calculate the loss.  
For the cross-entropy loss function, we can combine both the original loss and the new loss with a parameter $\lambda \in [0,1]$. The loss function from Eq.~\eqref{equ:standard_ce} is updated as follows.
\begin{equation}
    \mathcal{L} = \lambda \mathcal{L}_{ce}(\bm{p}, \bm{y}) + (1- \lambda) \mathcal{L}_{ce}(\bm{\hat{p}}, \bm{y}) \label{equ:lrpfinetune_ce}
\end{equation}
where $\mathcal{L}_{ce}$ denotes the cross-entropy loss and $\bm{y}$ is the ground truth label. We can also use $\bm{\hat{p}}$ for the SCST optimization and the reward formula from Eq.~\eqref{equ:standard_cider} is re-written as follows.
\begin{equation}
    R = \mathbbm{E}_{S^s, S^{greedy} \backsim \bm{\hat{p}}}[\text{metric}(S^s,S^{gt})-\text{metric}(S^{greedy}, S^{gt})] \label{equ:lrpfinetunecider}
\end{equation}
where we replace the original probability distribution $\bm{p}$ with the LRP-inference one $\bm{\hat{p}}$.
$R$ is the reward, $S^s$ is the sampled sentence, $S^{greedy}$ is the greedily sampled sentence, and $S^{gt}$ is the referenced caption.

Different from standard fine-tuning, LRP-IFT disentangles the contributions of the visual information, $R(\bm{c}_t)$, and the hidden state, $R(\bm{h}_t)$. It selects and fine-tunes the more related features rather than training all the features generally. 

To evaluate the performance of the LRP-IFT, we observe the mean average precision (mAP) of the frequent object words\footnote{The frequent object words of Flickr30K are the same as Section\ref{sec:rocauc}. The top-25 frequent object words of MSCOCO2017 datasets include: clock, kitchen, picture, water, food, pizza, grass, building, bus, sign,
bathroom, baseball, dog, room, cat, plate, train, field, tennis, person, table, street, woman, people, man.}. The motivation of LRP-IFT is to guide the model to make more grounded captions rather than thoroughly enumerate all objects within an image. Therefore, we do not use the recall and F1 score. 

\begin{table}[tb]
    \centering
    \scriptsize
    \caption{The mean average precision (mAP) of the predicted frequent object words. \emph{(ce)} denotes that the models are trained only with cross-entropy loss and the other models are optimized with SCST. \emph{BU} and \emph{CNN} denote bottom-up features and CNN features. Bold numbers indicate better results. Higher mAP means less object hallucination.} 
    \begin{tabular}{l r r r r r r}\hline
     dataset                  &  \multicolumn{2}{c}{Flickr30K} & \multicolumn{2}{c}{MSCOCO2017}\\ \hline
        mAP                         &baseline & LRP-IFT      & baseline        &LRP-IFT      \\
    Ada-LSTM-CNN              &52.95    & \textbf{54.47}   & 72.29           &\textbf{73.85}  \\ 
    Ada-LSTM-BU               &63.84    & \textbf{64.61}   & 78.57           &\textbf{80.55}  \\ 
    MH-FC-CNN                 &55.98    & \textbf{57.71}   & \textbf{73.74}  & 73.42  \\ 
    MH-FC-BU                  &64.46    & \textbf{64.98}   & \textbf{78.10}  & 77.71  \\  \hdashline
    Ada-LSTM-CNN (ce)    &58.53    & \textbf{60.80}   & 73.65           &\textbf{74.00}  \\ 
    Ada-LSTM-BU (ce)     &60.70    & \textbf{65.01}   & 79.06           &\textbf{79.80} \\ 
    MH-FC-CNN (ce)       &55.50    & \textbf{59.23}   & \textbf{77.15}  & 76.87  \\ 
    MH-FC-BU (ce)        &64.08    & \textbf{66.10}   & 81.02           & \textbf{81.16}\\\hline

    \end{tabular}    \label{tab:mAP}
\end{table}

Table \ref{tab:mAP} lists the mAP of the models with or without LRP-IFT. 
We implement the LRP-IFT on two sets of pre-trained models. The first set of models are from Table \ref{tab:model_performance} that are optimized with SCST optimization, and we refer to Eq.~\eqref{equ:lrpfinetunecider} to fine-tune the models for one epoch. The second set of models are trained only with cross-entropy loss, denoted as \emph{(ce)} in the table and we refer to Eq.~\eqref{equ:lrpfinetune_ce} with $\lambda=0.5$ to fine-tune the models for one epoch.
For the baseline models, we fine-tune the two sets of models with standard SCST optimization or cross-entropy loss with the same training hyperparameters. 
\begin{table*}[tb]
    \centering
    \scriptsize
    \caption{The performance of the Ada-LSTM model and the MH-FC model with or without LRP-IFT on the test set of Flickr30K and MSCOCO2017 datasets. L. denotes LRP-inference fine-tuned models. \emph{(ce)} denotes that the models are trained only with cross-entropy loss and the other models are further optimized with SCST. \emph{BU} and \emph{CNN} denote bottom-up features and CNN features. $F_B$: $F_{BERT}$, C: CIDEr, S: SPICE, R: ROUGE-L, M: METEOR.
    }

    \begin{tabular}{l r r r r r r r c c r r r r r}
        \hline
        dataset                  &  \multicolumn{5}{c}{Flickr30K} & & & \multicolumn{5}{c}{MSCOCO2017}\\ 
                         & $F_B$     & C      & S      & R      & M        & ~& ~ &$F_B$ & C   & S   & R & M \\ \hline
         Ada-LSTM-CNN    &90.6       & 51.1   & 13.9   & 46.4   & 20.0     & ~& ~ &91.7       & 107.1  & 19.5   & 54.2   & 26.1\\ 
         L.Ada-LSTM-CNN  &90.6       & 50.9   & 14.0   & 46.7   & 20.1     & ~& ~ &91.2       & 106.8  & 19.2   & 54.0   & 26.0\\ 
         Ada-LSTM-BU     &90.0       & 63.8   & 16.4   & 49.3   & 22.1     & ~& ~ &91.0       & 111.6  & 19.2   & 55.3   & 25.9\\ 
         L.Ada-LSTM-BU   &90.0       & 61.9   & 16.5   & 49.2   & 21.9     & ~& ~ &91.0       & 111.1  & 19.3   & 55.2   & 25.9\\ 
         MH-FC-CNN       &89.9       & 53.3   & 14.5   & 46.5   & 20.5     & ~& ~ &91.1       & 108.8  & 20.1   & 54.6   & 26.5\\ 
         L.MH-FC-CNN     &89.7       & 52.7   & 14.2   & 46.0   & 20.0     & ~& ~ &91.0       & 107.5  & 20.1   & 54.3   & 26.3\\ 
         MH-FC-BU        &90.1       & 63.5   & 17.1   & 49.4   & 22.5     & ~& ~ &91.3       & 120.9  & 21.8   & 56.6   & 28.1\\ 
         L.MH-FC-BU      &90.1       & 63.5   & 17.0   & 49.1   & 22.4     & ~& ~ &91.3       & 120.8  & 21.9   & 56.7   & 28.1\\ \hdashline
         
         Ada-LSTM-CNN(ce) &89.7      & 44.6   & 13.3   & 44.4   & 19.0     & ~& ~ &91.7       & 96.2   & 18.1   & 52.9   & 25.1\\ 
         L.Ada-LSTM-CNN(ce)&89.6     & 43.5   & 13.1   & 44.0   & 18.9     & ~& ~ &91.5       & 92.3   & 18.0   & 52.1   & 24.9\\ 
         Ada-LSTM-BU(ce) &90.0       & 53.3   & 15.6   & 47.3   & 21.2     & ~& ~ &91.9       & 107.4  & 19.8   & 54.9   & 26.9\\ 
         L.Ada-LSTM-BU(ce)&89.9      & 52.2   & 15.6   & 47.0   & 21.2     & ~& ~ &91.7       & 103.0  & 19.6   & 54.1   & 26.3\\ 
         MH-FC-CNN (ce)  &89.7       & 46.5   & 13.7   & 44.8   & 19.4     & ~& ~ &90.7       & 97.1   & 18.8   & 53.1   & 25.5\\ 
         L.MH-FC-CNN(ce) &89.6       & 47.3   & 14.1   & 45.6   & 19.8     & ~& ~ &90.7       & 97.2   & 18.8   & 53.0   & 25.4\\ 
         MH-FC-BU(ce)    &90.0       & 52.3   & 15.2   & 46.2   & 20.9     & ~& ~ & 91.8      & 105.8  & 19.9   & 54.7   & 26.7\\ 
         L.MH-FC-BU(ce)  &89.8       & 52.7   & 15.3   & 46.5   & 21.0     & ~& ~ & 91.8      & 105.7  & 19.9   & 54.6   & 26.6\\ 
         \hline
    \end{tabular}
    \label{tab:lrp-finetune-model_performance}
\end{table*}
\begin{figure*}[bt]
    \centering
    \includegraphics[width=0.9\textwidth]{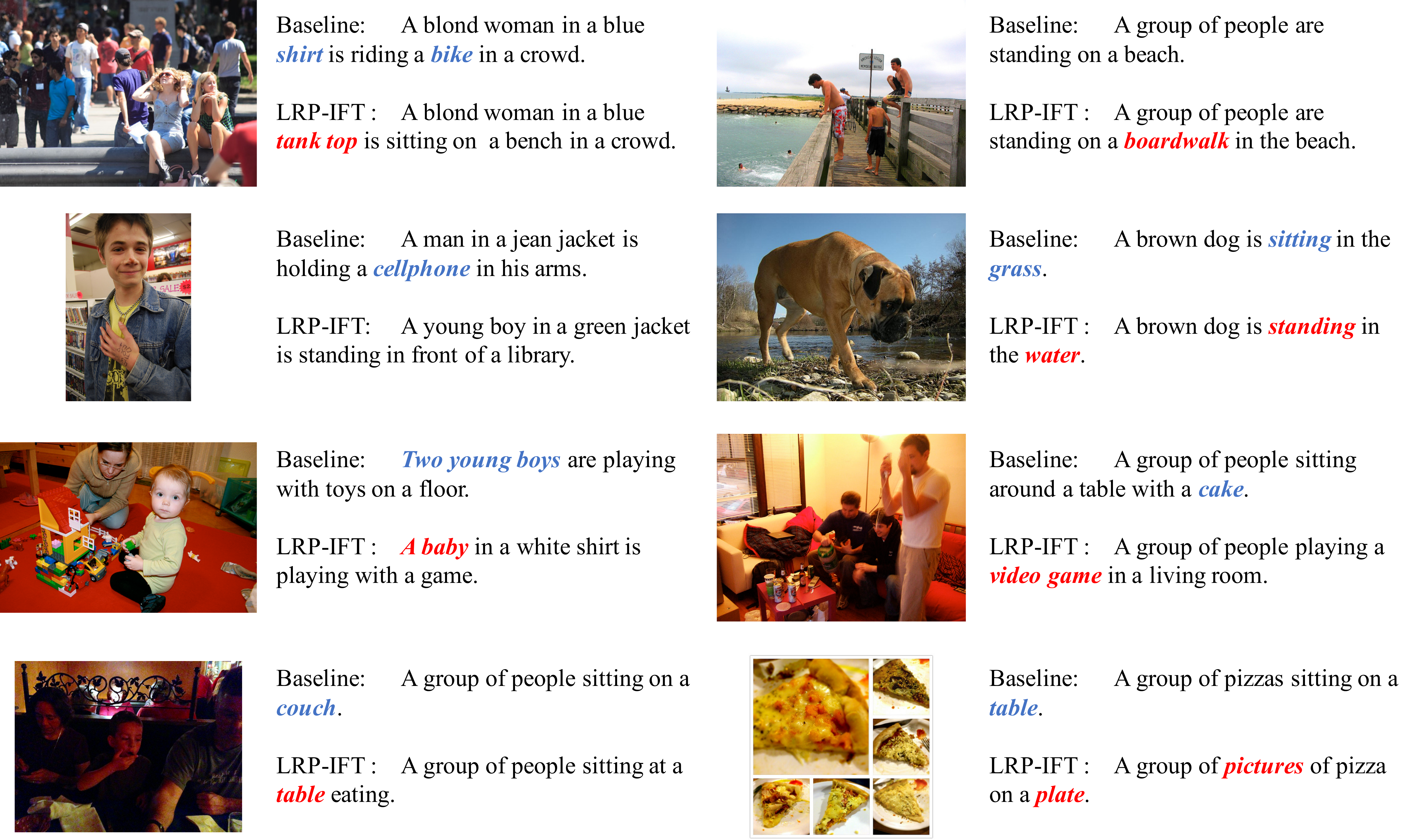}
    \caption{Examples of the captioning results with or without LRP-IFT. Blue color marks the hallucinated words and red color marks the words corrected by LRP-IFT.}
    \label{fig:example-captioning results}
\end{figure*}

As shown in Table \ref{tab:mAP}, the mAP is effectively improved after LRP-IFT for both sets of models except the MH-FC models trained on the MSCOCO2017 dataset. We discuss the mAP results from three aspects: 1) the MSCOCO2017 dataset has a more balanced vocabulary and more training data than the Flickr30K dataset, which results in less biased models. This also explains the more pronounced improvement of mAP on the Flickr30K dataset; 2) the multi-head attention mechanism has better grounding property as discussed in Section \ref{sec:localization}, which is the possible reason why LRP-IFT obtains similar mAP for the MH-FC model trained on the MSCOCO2017 dataset; 3) as expected, the image captioning models with bottom-up features consistently obtain higher mAP than those with CNN features, demonstrating the potential of better feature representation for visual-language models such as VIVO \cite{VIVO_NOC:hu2020vivo} and OSCAR \cite{OSCAR:li2020oscar}. 

Furthermore, LRP-IFT maintains the overall performance on the sentence level, as shown in Table \ref{tab:lrp-finetune-model_performance}. Figure \ref{fig:example-captioning results} illustrates some example captions of the baseline models and the LRP-inference fine-tuned models. LRP-IFT is conducted on the non-stop words and can improve the precision of the frequent object words. 
As shown in Figure \ref{fig:example-captioning results}, with LRP-IFT, the model can correct or remove the hallucinated words and maintain the sentence structure. 
This can partially explain why the sentence-level performance is very close to that of the baseline models. We will provide more detailed analyses of the sentence-level performance in Section \ref{sec:discussion}.

From the above analyses, the LRP-IFT can effectively de-bias and reduce object hallucination for a biased image captioning model, meanwhile, maintain the sentence-level performance in terms of $F_{BERT}$, CIDEr, SPICE, METEOR, and ROUGE-L. On the other hand, this fine-tuning strategy does not degrade the performance of a less biased image captioning model notably. 
We remark that the LRP-IFT requires no additional training parameters and human annotations. 
The fine-tuning procedure is also analogous to the human's recognition process that we first build prior knowledge by learning the objects, relations, and attributes and update related features when facing new shifts in distributions.

\begin{table*}[tb]
    \centering
    \scriptsize
    \caption{The average $c(S^{gt})$ over the ground truth captions from two sets of samples: the LRP-IFT-improved set, where LRP-IFT increases the CIDEr scores, and the LRP-IFT-degraded set, where LRP-IFT decreases the CIDEr scores. \emph{(ce)} denotes that the models are trained only with cross-entropy loss. The other models are further optimized with SCST. \emph{BU} and \emph{CNN} denote bottom-up features and CNN features. Bold numbers indicate lower counts of the ground truth words in the training set. This statistic can be interpreted as a heuristic for training data density.} 
    \begin{tabular}{l c c c c }\hline
      dataset                 &  \multicolumn{2}{c}{Flickr30K} & \multicolumn{2}{c}{MSCOCO2017}\\ \hline
    average counts            & LRP-IFT-improved        & LRP-IFT-degraded      & LRP-IFT-improved               & LRP-IFT-degraded     \\
    Ada-LSTM-CNN              &\textbf{26.1}   & 35.2          & \textbf{123.7}         & 134.0 \\ 
    Ada-LSTM-BU               &\textbf{30.1}   & 31.4          & \textbf{130.8}         & 134.7  \\ 
    MH-FC-CNN                 &\textbf{29.3}   & 31.4          & \textbf{124.4}         & 132.8  \\ 
    MH-FC-BU                  &\textbf{29.3}   & 29.7          & \textbf{118.7}         & 139.0 \\  \hdashline
    Ada-LSTM-CNN (ce)         &34.4   & \textbf{28.5}          & \textbf{124.4}         & 137.0 \\ 
    Ada-LSTM-BU (ce)          &31.7   & \textbf{28.1}          & \textbf{119.0}         & 150.6 \\ 
    MH-FC-CNN (ce)            &\textbf{29.4}   & 30.6          & \textbf{128.0}         & 142.6 \\ 
    MH-FC-BU (ce)             &\textbf{22.6}   & 35.9          & \textbf{124.7}         & 148.5 \\\hline

    \end{tabular}    \label{tab:frequency_rare_word_gt}
\end{table*}
\subsection{Discussion and outlook}
\label{sec:discussion}

In the experiments of LRP-IFT, we have observed that LRP-IFT alleviates the object hallucination issue of image captioning models measurably. However, we can also see that LRP-IFT does not effectively improve sentence-level performance. In this part, we will further analyze the effects of the LRP re-weighting mechanism and we will take a closer look at the samples where LRP-IFT improves the sentence-level performance. We conclude by proposing a potential future direction where the LRP-inference training can be helpful.

%

\subsubsection{On limitations of the LRP re-weighting mechanism}

We performed an analysis on the samples where LRP-IFT improves or degrades sentence-level performance. At first, for each word in a \emph{ground truth} caption, we computed the count of that word within the \emph{training set}. Then, for each ground truth caption in the test set, we find the minimum of the word counts, denoted as $c(S^{gt})$, over the \emph{non-stop} words in the caption $S^{gt}$:
\begin{equation}
    c(S^{gt})=\min_{w_t \in S^{gt}} count(w_t)
\end{equation}
where $count(w_t)$ returns the counts of the word $w_t$ in the training set. 

This statistic $c(S^{gt})$ for test set captions can be viewed as a heuristic $1$-gram estimate of the training data density for the linguistic modality of image captioning. For images with multiple ground truth captions, we take the minimum of $c(S^{gt})$ over all the captions of one image. We verified that taking the average yields the same qualitative results.

Finally, we compute the average of this heuristic $c(S^{gt})$ for two sets of images: 1) the images on which LRP-IFT improves the predictions compared to the baseline model and 2) the images for which LRP-IFT degrades the predictions compared to the baseline model. We refer to the sentence-level evaluation metrics, such as the CIDEr score, to separate the two sets of image samples.

Table \ref{tab:frequency_rare_word_gt} lists the results of average $c(S^{gt})$ using CIDEr scores for performance comparison. We observe a clear correlation across most of the models (except only one): The LRP-IFT-improved set exhibits a lower average $c(S^{gt})$, while the LRP-IFT-degraded set shows a higher average $c(S^{gt})$. In summary, LRP-IFT achieves a tradeoff. It performs worse on those test images with a higher estimate of the sample density, where the base model seemingly generalizes sufficiently well. On the other hand, it achieves an improvement on images with lower training data density. The results using other metric scores for comparison lead to the same finding. This makes intuitively sense as one can expect that captions supported by a higher amount of training data would profit less from learning with explanations. A similar correlation for using explanations to improve age prediction models using image data is reported in \cite{Weber2020Towards}. The authors observe that using explanations improves predictions on the poorly performing age subset 48-53 years, which has a small sample size, while slightly degrades the performance on age subsets with larger sample sizes.

There are further possible reasons for the non-improved sentence-level performance. \cite{OBJECTHALLU:rohrbach2018object} points out that hallucinating less does not necessarily render higher sentence-level evaluation metrics \cite{GenderBias:hendricks2018women, HINT:selvaraju2019taking, MitigateGender:tang2020mitigating}, which is also in line with our observations in Table \ref{tab:lrp-finetune-model_performance}. Furthermore, LRP-IFT implements the re-weighting mechanism on top of pre-trained models as a fine-tuning step, making it challenging to achieve larger changes over pre-trained models.

\subsubsection{An outlook for re-weighting mechanisms based on explanations}
Based on the above analyses, we surmise that the LRP re-weighting mechanism could be helpful for novel object captioning (NOC). NOC aims to predict those object words that are unseen by the model during training. It also faces the challenge of unbalanced training data, in an even more extreme case where some object words are not shown in the training data.
For example, \cite{POINGTING_NOC:li2019pointing} proposed a pointing mechanism to combine the sentence correlation representation and object representation, which dynamically decides whether to include an object word from a detection model. 
The LRP re-weighting mechanism could be helpful here to better guide the model when and where to include the detected objects in the caption.

\section{Conclusion}
\label{sec:conculsion}
We adapt LRP and gradient-based explanation methods to explain the attention-guided image captioning models beyond visualizing attention. With extensive qualitative and quantitative experiments, we demonstrate that explanation methods provide more interpretable information than attention, disentangle the contributions of the visual and linguistic information, help to debug the image captioning models such as mining the reasons for the hallucination problem. With the properties of LRP explanations, we propose an LRP-inference fine-tuning strategy that can successfully de-bias  image captioning models and alleviate object hallucination. The proposed fine-tuning strategy requires no additional annotations and training parameters.

\section*{Acknowledgements}
This work was supported by the Ministry of Education of Singapore (MoE) Tier 2 grant MOE2016-T2-2-154.
This work was also partly supported by the German Ministry for Education and Research as BIFOLD (ref.\ 01IS18025A and ref.\ 01IS18037A),
and TraMeExCo (ref.\ 01IS18056A), by the European Union’s Horizon 2020 programme (grant no.\ 965221) and by the Research Council of Norway, via the SFI Visual Intelligence, project number 309439.

\bibliographystyle{IEEEtran}
\bibliography{ms}

\end{document}